\documentclass[runningheads]{llncs}
\usepackage{graphicx}

\usepackage{tikz}
\usepackage{comment} 
\usepackage{amsmath,amssymb} 
\usepackage{xcolor}
\usepackage{colortbl}

\usepackage{tikz}
\usepackage{epsfig}
\usepackage{multirow}
\usepackage{caption}
\usepackage{subcaption}
\usepackage{comment}
\usepackage{wrapfig}
\usepackage{bm}
\usepackage[pagebackref=true,breaklinks=true,letterpaper=true,colorlinks,bookmarks=false]{hyperref}
\newcommand{\etal}{\emph{et al.}}
\newcommand{\eg}{\emph{e.g.}}
\newcommand{\ie}{\emph{i.e.}}
\newcommand{\wrt}{\emph{w.r.t.}}
\newcommand{\etc}{\emph{etc}.}

\begin{document}
\pagestyle{headings}
\mainmatter
\def\ECCVSubNumber{3199}  

\title{Inter-Image Communication for \\ Weakly Supervised Localization} 

\titlerunning{$I^2C$}
%
\author{Xiaolin Zhang \and
Yunchao Wei \and
Yi Yang}
\authorrunning{Xiaolin Zhang, Yunchao Wei, and Yi Yang}
%
\institute{ReLER, AAII, University of Technology Sydney\\
\email{\{Xiaolin.Zhang-3@student.,Yunchao.Wei@,Yi.Yang@\}uts.edu.au}}

\maketitle

\begin{abstract}
   Weakly supervised localization aims at finding target object regions using only image-level supervision. 
   However, localization maps extracted from classification networks are often not accurate due to the lack of fine pixel-level supervision.
  In this paper, we propose to leverage pixel-level similarities across different objects for learning more accurate object locations in a complementary way. 
   Particularly, two kinds of constraints are proposed to prompt the consistency of object features within the same categories.    
   The first constraint is to learn the stochastic feature consistency among discriminative pixels that are randomly sampled from different images within a batch.
   The discriminative information embedded in one image can be leveraged to benefit its counterpart with inter-image communication.
   The second constraint is to learn the global consistency of object features throughout the entire dataset.
  We learn a feature center for each category and realize the global feature consistency by forcing the object features to approach class-specific centers.
   The global centers are actively updated with the training process.
   The two constraints can benefit each other to learn consistent pixel-level features within the same categories, and finally improve the quality of localization maps.
   We conduct extensive experiments on two popular benchmarks,~\ie, ILSVRC and CUB-200-2011. 
   Our method achieves the Top-1 localization error rate of $45.17\%$ on the ILSVRC validation set, surpassing the current state-of-the-art method by a large margin. 
   The code is available at \url{https://github.com/xiaomengyc/I2C}.

\end{abstract}
\section{Introduction}\label{sec:introduction}

Deep learning has achieved great success on various tasks, \eg, classification~\cite{simonyan2014very,szegedy2016rethinking}, detection~\cite{girshick15fastrcnn,ren2015faster}, segmentation~\cite{chen2014semantic,chen2018encoder,zhao2017pyramid,huang2020ccnet,cheng2019spgnet}~\etal.
\begin{figure}[thbp]
  \centering
  \includegraphics[width=0.7\textwidth]{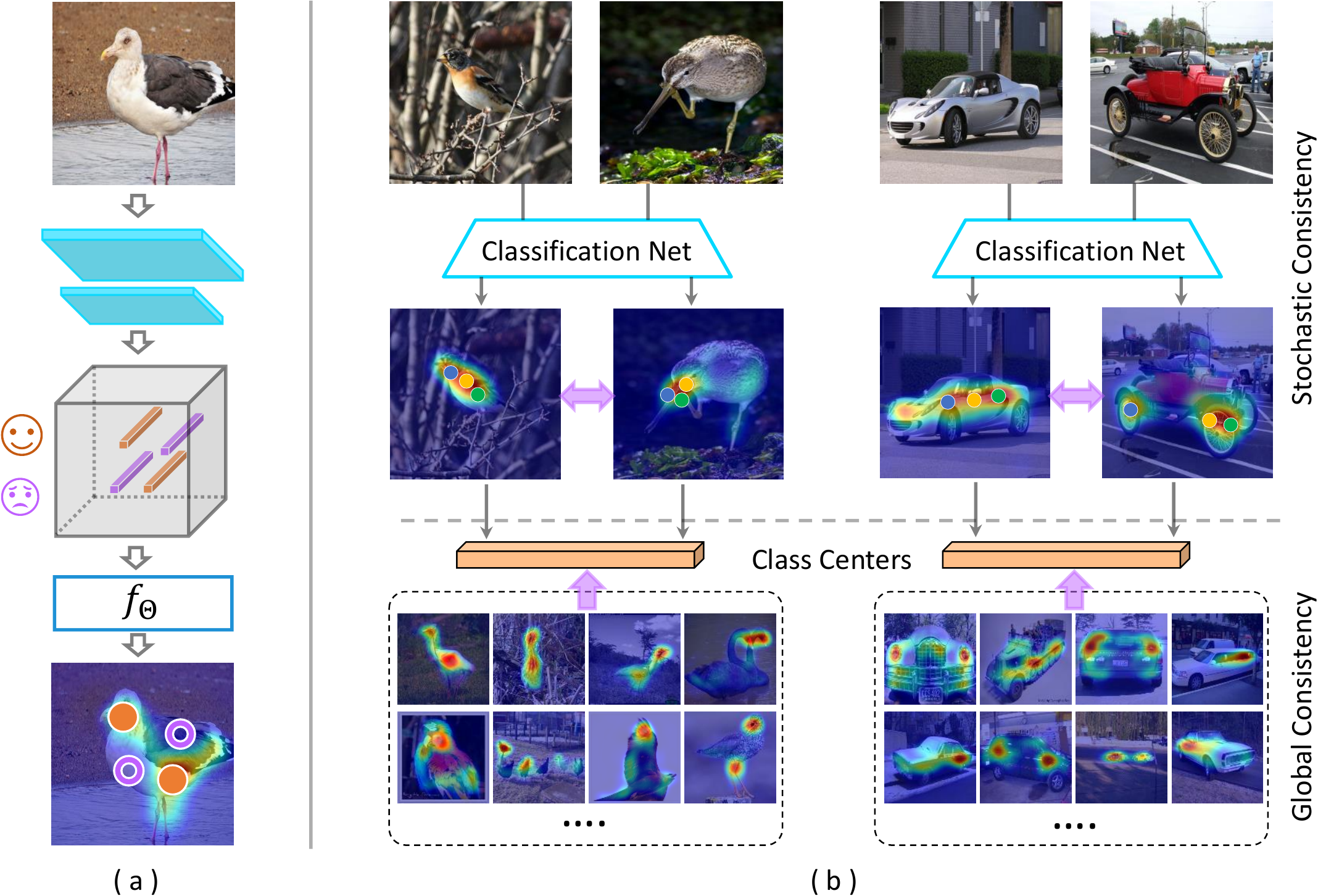}
  \caption{
  (a) Convolutional operations preserve the relative pixel positions. Inconsistent response scores of different pixels on the class activation map are essentially caused by the inconsistent learned features.
  (b) The proposed Stochastic Consistency (SC) and Global Consistency (GC) are to align the object-related feature vectors. \textit{Upper:} Pixel features of images sampled from the same category can reach consistency between the highly confident points within the same mini-batch. \textit{Bottom:} We actively learn class-specific centers and push object features of the same category across different mini-batch towards the centers.
  }\label{fig0}
\end{figure}
In this paper, we focus on the Weakly Supervised Object Localization (WSOL) problem.
Briefly, WSOL tries to locate object regions within given images using only image-level labels as supervision.
Currently, the standard practice for this task is to train a  convolutional classification network supervised by the given image-level labels.
The convolutional operations can preserve relative positions of the input pixels so that activations from high-level layers can roughly indicate the position of target objects.
Some previous works~\cite{zhou2015cnnlocalization,zhang2018adversarial,zhang2018self} have already explored how to produce object localization maps effectively.
These methods obtain localization maps by aggregating feature maps with a fully connected layer.
Figure~\ref{fig0}\textcolor{red}{a} shows the general pipeline for generating localization maps.
Given an image of a bird, it is firstly fed into some convolution layers to yield the feature maps. 
These feature maps are then processed by a function $f_\Theta$ to get the localization maps of the right class using the methods in CAM~\cite{zhou2015cnnlocalization}, ACoL~\cite{zhang2018adversarial}, ADL~\cite{Choe_2019_CVPR}, \etc. 
Ideally, the localization maps are expected to highlight all the object regions and depress the background.
Unfortunately, only some sparse parts are highlighted and cannot cover the entire target objects, which is a major problem in practice.
We try to alleviate this issue based on the following intuition.
Theoretically, for a specific output score of a pixel, the input features of the function $f_\Theta$ may be various.
However, there are two observations in practice,~\ie, 1) convolution operations preserve the relative positions between the input and output feature maps~\cite{zhou2015cnnlocalization,zhang2018adversarial}; 2) features of the same category objects tend to lie in the same clusters and similar input features produce similar output values~\cite{zhou2018interpreting}. 
Therefore, the root cause for this problem is that the network fails to learn consistent feature representations for pixels belonging to the object of interest.
In Figure~\ref{fig0}\textcolor{red}{a}, the low response scores in the localization maps (\textcolor{pink}{pink} circles) of the bird are caused by the low-quality or non-discriminative features in the intermediate feature maps, while the high response scores (\textcolor{orange}{orange} dots) are produced by the decent and discriminative features.
Several efforts have been taken to alleviate this issue.
MDC~\cite{wei2018revisiting} attempt to learn consistent features of different parts by enlarging receptive fields with dilated convolutions~\cite{chen2017deeplab}, but it is prone to draw into background noises.
SPG~\cite{zhang2018self} employs an auxiliary loss to enforce the consistency of object features using self-produced pseudo-masks as supervision.
Nevertheless, the both methods only consider inter-pixel correlations within an image, and involve sophisticated network modules and many extra computational resources.

In this paper, we alleviate this issue by employing two constraints on the object pixels across images with the cost of negligible extra resources.
Different from MDC and SPG, we argue that not only pixels within a object should keep close, but more importantly, object pixels of different images in the same category should also semantically keep consistent in the high-level feature space.
In Figure~\ref{fig0}\textcolor{red}{b}, different parts  of the same category objects~\eg, heads and bodies of the birds, the wheels and bodies of the vehicles, are highlighted in different images.
The corresponding features of these highlighted pixels do not necessarily very close, but we force them to communicate with each other to learn more consistent and robust features, and thereby, produce better localization maps.
We propose to realize the pixel-level communications by employing two constraints~\ie, Stochastic Consistency (SC) and Global Consistency (GC).
The two constraints act as auxiliary loss functions to train classification networks.
With the training process, convolution networks are not only looking for discriminative patterns to support the classification purpose, but communicating between different object patterns for learning more consistent and robust features.
Consequently, more accurate object regions will be discovered as we desired.
Specifically, the proposed SC is to constrain the object features of the same category within a batch.
We firstly feed training images through a classification network for obtaining localization maps.
Then, we select several confident seed points among the pixels with high scores in the maps. 
Finally, features of these seed pixels can communicate with each other across images within the same category.
We attain the inter-image communication by optimizing the Euclidean distance between high-level features of the seed pixels.
In Figure~\ref{fig0}\textcolor{red}{b}, given two images of the same class, different parts of the target objects are highlighted. 
Object features can reach consistencies by narrowing the distances between the seed points of different images.
Notably, due to the lack of pixel annotations in our task, we firstly ascertain a portion of object pixels for each image.
Pixels with high scores in localization maps usually lie in object regions with high confidence~\cite{zhang2018adversarial,zhang2018self}.
Meanwhile, it is also significant to avoid the influence from the background by only considering the confident object regions. 
Then, inter-image pixel-level communication can be explicitly accomplished.

Due to the limitation of the Stochastic Gradient Descent (SGD) optimization methodology,
SC can only keep the semantic consistency within a batch.
It cannot guarantee the class-specific consistency of the entire dataset.
To tackle this issue, we further introduce the Global Consistency (GC) to augment SC by constraining images across the entire training set.
As in Figure~\ref{fig0}\textcolor{red}{b}, image features are forced towards their class-specific global centers.
In detail, the class-specific centers are actively maintained.
We adopt a momentum strategy to update the class centers.
During each training step, class-specific centers are updated using the seed features and memory centers.
The ultimate goal of GC is to push object features approaching their global centers throughout the training set.
GC constrains the object features across images.
It is noticeable that the proposed SC constraint does not bring any extra parameters.
GC brings a few parameters which is negligible compared to the backbone parameters.

We name the proposed method as Inter-Image Communication ($I^2C$) model. 
We conduct extensive experiments on the ILSVRC~\cite{2009-imagenet}~\cite{ILSVRC15} and CUB-200-2011~\cite{WahCUB_200_2011} datasets. 
Our main contributions are three-fold:
\begin{itemize}
    \item{We propose to employ inter-image communication of objects in the same category for learning more robust and reliable localization maps under the supervision of image-level annotations.}
    \item{We propose two constraints.
    Stochastic consistency can keep the features of the same semantic within a batch close in the high-level feature space.  Global consistency can learn a class-specific center for each category and keep the features of the same category close throughout the training set.
    }
    \item{This work achieves a new state-of-the-art with the localization error rate of Top-1 $45.17\%$ on the ILSVRC~\cite{2009-imagenet,ILSVRC15} validation set and $44.01\%$ on the CUB-200-2011~\cite{WahCUB_200_2011} test set.} 
    
\end{itemize}

\section{Related Work}
\textbf{Weakly supervised segmentation} has a strong connection with our WSOL task~\cite{wei2015stc,wei2016learning,wei2017object,kolesnikov2016seed,2015-papandreou-weakly,wei2018revisiting,hou2018self,jiang2019integral,ahn2018learning,ahn2019weakly}.
It normally and firstly applies similar techniques to obtain pseudo masks, and then trains segmentation networks for predicting accurate masks.
DD-Net~\cite{shimoda2016distinct} studies the method of generating pseudo masks from localization maps, and proposes to improve the accuracy by removing noises via self-supervised learning.
OAA~\cite{jiang2019integral} accumulates localization maps with respect to different training epochs with the optimization process.
ICD~\cite{Fan_2020_CVPR} utilizes an intra-class discriminator to learn better boundaries between classes.
SEAM~\cite{Wang_2020_CVPR} proposes to use consistency regularization on predicted activation maps of various transforms as self-supervision for network learning.

\noindent\textbf{Weakly supervised detection and localization}
aims to apply an alternative cheaper way by only using image-level supervision~\cite{singh2017hide,liang2015towards,oquab2015object,durand2017wildcat,zhu2017soft}.
Recently, ADL~\cite{Choe_2019_CVPR} borrows the adversarial erasing idea from ACoL~\cite{zhang2018adversarial} to provide a more neat and powerful approach without bringing much more parameters and computational resources.
CutMix~\cite{yun2019cutmix} also explores the techniques of erasing patches of images. In addition to the erasing operation, CutMix mixes ground truth labels along with image patches.
SEM~\cite{zhang2020rethinking} and ~\cite{choe2020evaluating} reformulate the evaluation of WSOL problem. 
SEM also proposes an enhancement alternative approach to produce high-quality localization maps.
\section{Methodology}
\begin{figure*}[!t]
  \centering
  \includegraphics[width=0.8\textwidth]{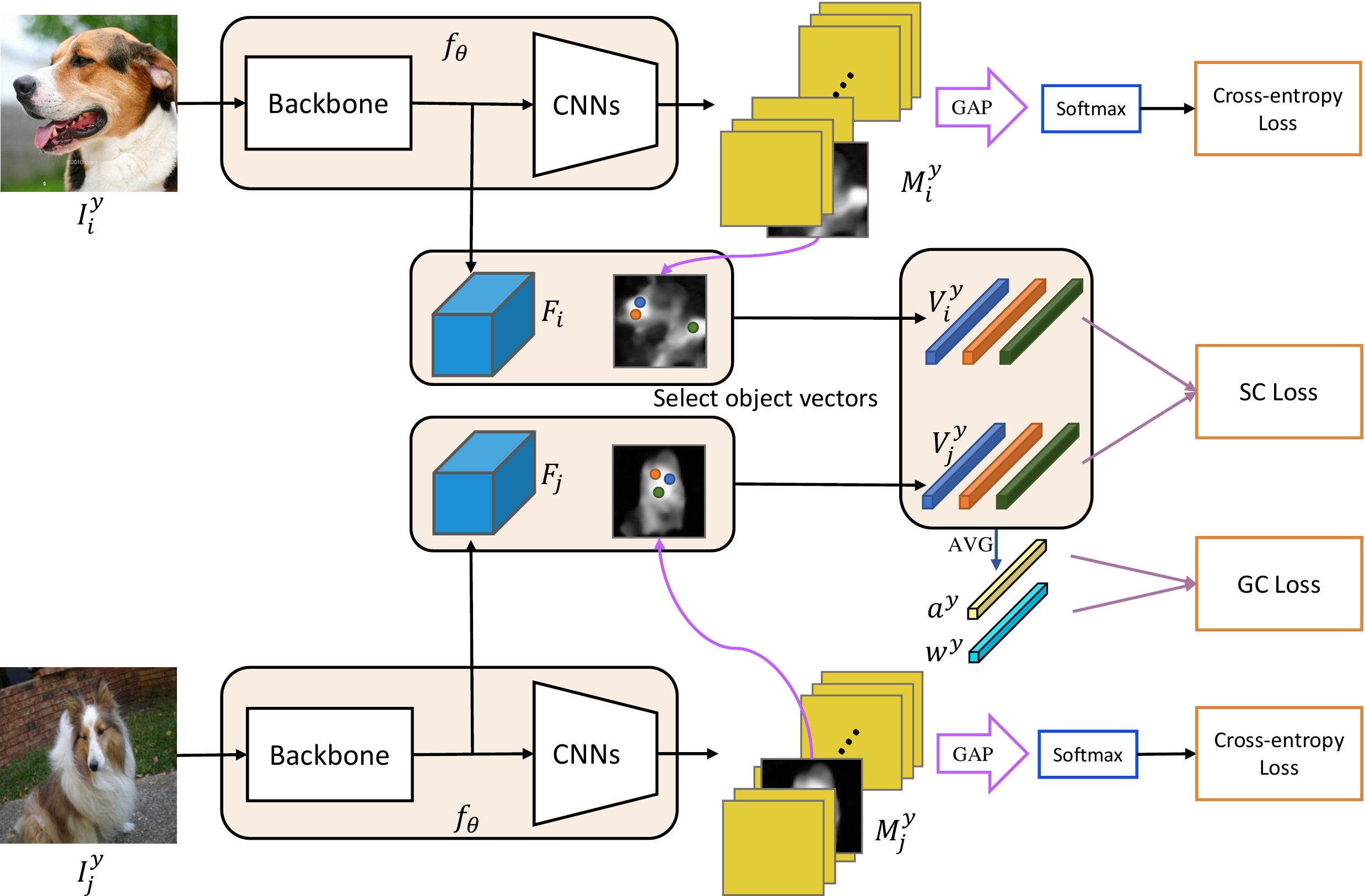}
  \caption{The structure of the proposed approach. Given a pair of images $(I_i^y,I_j^y)$ of the same category $y$, the localization maps $(M_i^y, M_j^y)$ can be obtained by forwarding them through the classification network $f_\theta$. Object seed vectors $(V_i^y, V_j^y)$ are extracted from the high-level feature maps $(F_i, F_j)$ according to the confident regions in the maps. Finally, the SC loss is employed on the object seed vectors. Also, the GC loss is employed on the averaged object feature $a^y$ from a batch and the global class-specific center $w^y$. GAP refers to Global Average Pooling. AVG refers to the average operation. }\label{fig-framework}
\end{figure*}

Figure~\ref{fig-framework} depicts the framework of the proposed approach.
Given a pair of images $(I_i^y, I_j^y)$ sharing the same category $y$, we firstly forward them to obtain high-level feature maps $(F_i, F_j)$ and localization maps $(M_i^y, M_j^y)$.
Then, we identify the reliable object regions according to the response scores in the produced localization maps.
We randomly sample $K$ representative seeds from the object regions. 
$K$ object vectors of the seeds, denoted as $(V_i^y, V_j^y)$, are extracted from the high-level feature maps $(F_i, F_j)$ for each image according to the spatial localization.
Next, we optimize the similarity between different object seed vectors using the Stochastic Consistency (SC) loss across images.
Additionally, the Global Consistency (GC) loss is also employed to encourage the object representative vector of each minibatch to approach the global class-specific vector $w^y$ throughout the training set.
We will describe the details of constructing the object seed vectors, the SC and GC losses in the following sections.

\subsection{Object Seed Vectors}\label{obj_representation}
Object seeds serve as the bridge for inter-image communication to narrow the distance between object pixels of the same category.
Our target is to consistently high-light object pixels in localization maps, which is equivalent to get consistent high-level object features as the analysis in Section~\ref{sec:introduction}.
The first obstacle we are facing is the lack of object pixel cues.
Some previous works~\cite{wei2017object,zhang2018adversarial,zhang2018self} employ a trade-off strategy of mining the confident object regions according to the scores in localization maps.
We further randomly select some seed pixels as the object representation, and optimize the distance of the seeds across images. 
Such that intact object regions are high-lighted consistently, and robustness can be improved by exchanging information with different objects.

In detail, given an input image $I\in \mathbb{R}^{3\times W_0 \times H_0}$ and its class label $y\in \{0,1,...,Y-1\}$,
we pick out the localization map from the output of the last convolution layer corresponding to the category $y$ following the simplified method in~\cite{zhang2018adversarial},
where $W_0$ and $H_0$ denote the width and height of the image, $Y$ is the number of total categories.
Suppose the normalized map is $M^y\in [0,1]^{W\times H}$ and the map before normalization is $M^{*y}\in \mathbb{R}^{W\times H}$, where $W$ and $H$ are the width and height of the localization map, we calculate the normalized map by following $M^y_{i,j}=\frac{M^{*y}_{i,j}-\min(M^{*y})}{\max(M^{*y}) - \min(M^{*y})}$, where $i$ and $j$ are the indices of the map. 
For each image in a batch, we extract $K \in [1, W\times H]$ object-related seeds within the object regions.
The object regions can be identified according to the scores in the localization maps.
To be specific, pixels whose corresponding values in localization maps are higher than a threshold $\delta \in (0,1)$ are considered as reliable object regions.
We randomly select $K$ pixels from the object regions as the object representation seeds of the image $I$.
We denote the object seed vectors as $V \in \mathbb{R}^{K\times D}$ which is extracted from high-level feature maps, where $D$ is the dimension of the vectors,~\ie, the number of the channels in the feature maps.
In this way, we finally obtain $K$ vectors of the most discriminative object regions.
Next, these vectors are employed to communicate with the other images in the same category.

\subsection{Stochastic Consistency}

Features of pixels belonging to the same category should be close.
Although classification loss can drive networks to find the most discriminative patterns to produce correct classification scores, the learned features of the same objects are not necessarily similar due to the lack of pixel-level supervision.
In other words, features between different pixels of objects are not consistent, which deviates from the requirement of highlighting integral object regions. 
We argue that more accurate localization maps can be obtained by keeping the consistency between features of objects in the same class.
Object features of different images can communicate in a complementary approach, and thus the entire object regions can be consistently highlighted.
We propose a Stochastic Constraint (SC) to drive the consistency of pixels from different objects of the same category in a minibatch.
Given a batch $B=\{(I_i, y_i)|i=0,1,...,N^B\}$ of randomly sampled images,  we can find any two images $I_i$ and $I_j$ whose class labels are the same, namely, $y_i = y_j$, where $N^B$ is the batch size.
After forwarding them through the network, we obtain the object seed vectors according to Section~\ref{obj_representation}.
We denote the object seed features as $V_{i}\in \mathbb{R}^{K\times D}$ and $V_{j}\in \mathbb{R}^{K\times D}$, respectively.
We expect the two images can communicate by explicitly constraining the pixels belonging to the same category.
Particularly, we propose a constraint to optimize the L2-norm between the object features of $I_i$ and $I_j$ according to Eq.~\eqref{eq_local_loss}.
\begin{equation}\label{eq_local_loss}
    L_{sc} = \frac{1}{K} \sum_{k=0}^{K-1}||v^k_{i} - v^k_{j}||_2^2,
\end{equation}
where $K$ is the number of randomly selected seeds, 
and $v^k\in \mathbb{R}^{D}$ is the $k_{th}$ row of $V$.
Thus, object features can communicate with each other across images in a batch, and influences from the background is not involved.

\subsection{Global Consistency}
Deep networks are trained with the SGD based optimization algorithms,~\eg, Adam~\cite{kingma2014adam}, Adagrad~\cite{duchi2011adaptive} and RMSprop~\cite{graves2013generating}.
These methods construct minibatches by randomly sampling images  to perform training steps, which means SC can only constrain images within the same minibatch.
To overcome this limitation, we propose to learn a global feature center for each category, so that features extracted from a batch can be optimized to approach the class-specific global centers.
The object features of each class are hence gradually consistent with the global vectors.

We maintain a memory bank $W\in \mathbb{R}^{Y\times D}$.
The $y_{th}$ row of $W$ denoted as $w_y$ is the global center of category $y$.
For each batch, we obtain one representation vector for each category by averaging the object seed vectors sharing the same class labels. 
Formally, we denote $B^{y}=\{(I_i, y_i)|y_i=y\}$ and $\{V^y_k|k=0,1,..., K|B^y|\}$ as the class-specific subset and the object seed vectors corresponding to the category $y$ in a batch , where $|B^y|$ is the image number in the subset $B^y$.
We extract one representation vector $a^y$ for category $y$ from every minibatch.
Formally, the representation vector $a^y$ of class $y$ in a batch is obtained according to Eq.~\eqref{eq_batch_represention}.
\begin{equation}\label{eq_batch_represention}
    a^y = \frac{1}{K|B^y|}\sum_{k=0}^{K|B^y|-1} V^y_k.
\end{equation}

Global vector $w_y$ \wrt ~class $y$ in a minibatch is updated during each training step.
We do \textit{not} update the memory bank during the backward process.
We propose a simple yet effective updating procedure for learning the memory matrix.
For the class $y$, its global representation $w_y$ is as in Eq.~\eqref{eq_global_update}.
\begin{equation}\label{eq_global_update}
    w^y = (1-\eta_{t^y}^y)w^y + \eta_{t^y}^y a^y, 0<\eta_{t^y}^y<1,
\end{equation}
where $\eta_{t^c}^c$ is the updating rate of the class $y$ at step $t^y$.
We adopt a class-specific updating rate as given in Eq.~\eqref{eq_lr_decay} for learning the global centers.
\begin{equation}\label{eq_lr_decay}
    \eta_{t^y}^y =  e^{-\alpha t^y}, 0<\alpha<1,
\end{equation}
where $t^y$ counts the updating steps of class $y$, and $\eta_{t^y}^y$ is the update rate of the class $y$ at learning step $t^y$.
Note the updating step $t^y$ is also class-specific and maintained throughout the training process, because the updated counters of different categories are not the same due to the random sampling procedure of each batch.
$\alpha$ is a hyper-parameter for decaying the learning rate of the global representation $w^y$.
Thereby, $w^y$ will be gradually stable with the training process.

The features of each batch can be optimized to approach the global representation as in Eq~\eqref{eq_loss_global}.
\begin{equation}\label{eq_loss_global}
    L_{gc} = \frac{1}{|Y^B|} \sum_{y \in Y^B}||a^{y} - w^{y}||_2^2,
\end{equation}
where $Y^B$ is the label set of the mini-batch.

We further apply a typical cross-entropy loss for classification and denote it as $L_{cls}$.
In total, the optimization of our approach is a joint training process with three items following Eq.~\eqref{eq_total_loss}
\begin{equation}\label{eq_total_loss}
    L = L_{cls} + \lambda_{1} L_{sc} + \lambda_{2} L_{gc},
\end{equation}
where $\lambda_1$ and $\lambda_2$ are for trading-off the three loss items.

In \textit{testing}, we simply feed testing images into the network, and obtain the localization maps~\wrt ~the predicted class labels.
The extracted localization map is then normalized and resized to the original size of the input image by the bilinear interpolation.
Following the baseline methods~\cite{zhang2018adversarial,zhang2018self,zhou2015cnnlocalization}, we leverage the same strategy in CAM~\cite{zhou2015cnnlocalization} for generating the bounding boxes of the target objects.
Specifically, we firstly binarize the localization maps by a threshold for separating the object regions from the background.
Afterward, we draw tight bounding boxes that can cover the largest connected area of the foreground pixels. The thresholds for splitting the object regions are adjusted to the optimal values.
For more details, please refer to~\cite{zhou2015cnnlocalization}.
\section{Experiments}
\subsection{Experiment setup}\label{exper-setup}
\textbf{Datasets}
We evaluate the proposed method following the previous methods, \eg, CAM~\cite{zhou2015cnnlocalization}, ACoL~\cite{zhang2018adversarial}, SPG~\cite{zhang2018self} and ADL~\cite{Choe_2019_CVPR}.
Two datasets, \ie, ILSVRC~\cite{2009-imagenet,ILSVRC15} and CUB-200-2011~\cite{WahCUB_200_2011} are applied to train classification networks for a fair comparison with the baselines. 
ILSVRC is a widely recognized large-scale classification dataset including 1.2 million images of 1,000 categories for training and 50,000 images for validation. 
Images in both the training and validation sets are well annotated with image categories and tight  bounding boxes of objects.
CUB-200-2011~\cite{WahCUB_200_2011} includes 11,788 images from 200 different species of birds. 
It is splitted into the training set of 5,994 images and the testing set of 5,794 images. 
Similarly, all images are annotated with class labels and tight bounding boxes.
In our experiments, the proposed method is learned with the training sets using only image-level labels as supervision. 
The annotated bounding boxes on the validation set of ILSVRC and the testing set of CUB-200-2011 are employed for the evaluation.

\textbf{Evaluation metrics}
We apply the recommended metric in~\cite{ILSVRC15} to evaluate the localization maps following the baseline algorithms~\cite{zhou2015cnnlocalization,zhang2018adversarial,zhang2018self,Choe_2019_CVPR}.
To be specific, it calculates the percentage of the images that satisfy the following two conditions.
First, the predicted classification labels match the ground-truth categories. 
Second, the predicted bounding boxes have over 50\% IoU with at least one of the ground-truth boxes. 
In order to have a more explicit and pure comparison in localization ability, we calculate and compare the localization accuracy given ground-truth labels. 
We denote these results as the Gt-known localization accuracy.
This Gt-known localization accuracy removes the influence of classification results and is much fairer in comparing the localization ability.

\textbf{Implementation details}
We implement the proposed algorithm based on three popular backbone networks,~\eg VGG16~\cite{simonyan2014very}, ResNet50~\cite{he2016deep} and InceptionV3~\cite{szegedy2016rethinking}.
We make the same modifications on the networks to obtain localization maps following ACoL~\cite{zhang2018adversarial} and SPG~\cite{zhang2018self}.
We use the simplified method in ACoL~\cite{zhang2018adversarial} to produce localization maps, while the erasing branch is not applied.
We enable the proposed constraints after finetuning the parameters for a few epoches to obtain a good initialization of the newly added parameters.
In the ablation experiments, we compare a plain version network without SC and GC for comparison, named InceptionV3-plain. 
In order to assure that each batch contains images of the same category, we randomly draw $40$ categories and then randomly sample two images from the selected categories, constructing the image batch size of $80$.
We apply multiple hyper-parameters \ie, $\lambda_1$, $\lambda_2$, $\delta$, $\alpha$ and $K$, and conduct extensive experiments for studying the impact of these variables.
The global memory centers are randomly initialized.
We set $\delta=0.7$ following SPG~\cite{zhang2018self}.

\subsection{Comparison with the state-of-the-arts}
\newcolumntype{g}{>{\columncolor[gray]{0.9}}c}
\begin{table*}[t]\setlength{\tabcolsep}{4pt}
  \small
  \centering
  \begin{tabular}{l|c|ccc|cc}
    \hline
    \hline
     \multirow{2}{*}{Methods} & \multirow{2}{*}{Backbone} & \multicolumn{3}{c}{Loc Err.} & \multicolumn{2}{|c}{Cls Err.}\\
     \cline{3-7}
      & & Top-1 & Top-5 & Gt-known & Top-1 & Top-5 \\
    \hline
     CAM~\cite{zhou2015cnnlocalization} & AlexNet~\cite{krizhevsky2012imagenet} & 67.19 & 52.16 & 45.01 & 42.6 & 19.5 \\
     \hline
     CAM~\cite{zhou2015cnnlocalization} & GoogLeNet~\cite{szegedy2015going} & 56.40 & 43.00 & 41.34 & 31.9 & 11.3\\
     HaS-32~\cite{singh2017hide} &GoogLeNet~\cite{szegedy2015going}& 54.53 & - & 39.43 & 32.5 & -  \\
     ACoL~\cite{zhang2018adversarial} & GoogLeNet~\cite{szegedy2015going} & 53.28 & 42.58 & -  & 29.0 & 11.8\\
     DANet~\cite{xue2019danet} & GoogLeNet~\cite{simonyan2014very} & 52.47 & 41.72 & - & 27.5 & 8.6\\
     \hline
     Backprop~\cite{simonyan2013deep}  & VGG16~\cite{simonyan2014very} & 61.12 & 51.46 & - & - \\
     CAM~\cite{zhou2015cnnlocalization} & VGG16~\cite{simonyan2014very}  & 57.20 & 45.14 & - &  31.2 & 11.4 \\
     CutMix~\cite{yun2019cutmix} & VGG16~\cite{simonyan2014very} & 56.55 & - & - & - & - \\
     ADL~\cite{Choe_2019_CVPR} & VGG16~\cite{simonyan2014very} & 55.08 & - & - & 32.2  &  -\\
     ACoL~\cite{zhang2018adversarial} & VGG16~\cite{simonyan2014very} & 54.17 & \textbf{40.57} & 37.04 & 32.5 & 12.0\\
     \rowcolor[gray]{0.9}
     $I^2C$-Ours & \textcolor{black}{ VGG16}~\cite{simonyan2014very} & \textcolor{black}{\textbf{52.59}} & \textcolor{black}{41.49} & \textcolor{black}{\textbf{36.10}}  & \textcolor{black}{30.6} & \textcolor{black}{10.7}\\
     \hline
     CAM~\cite{zhou2015cnnlocalization} & ResNet50-SE~\cite{he2016deep,hu2018squeeze} & 53.81 & - & - & 23.44  &  -\\
     CutMix~\cite{yun2019cutmix} & ResNet50~\cite{he2016deep} & 52.75 & - & - & 21.4  &  5.92\\
     ADL~\cite{Choe_2019_CVPR} & ResNet50-SE~\cite{he2016deep,hu2018squeeze} & 51.47 & - & - & 24.15  &  -\\
     \rowcolor[gray]{0.9}
     $I^2C$-Ours & \textcolor{black}{ResNet50}~\cite{he2016deep} & \textcolor{black}{\textbf{45.17}} & \textcolor{black}{\textbf{35.40}} & \textcolor{black}{\textbf{31.50}}  & \textcolor{black}{23.3} & \textcolor{black}{6.9} \\
     \hline
     CAM~\cite{zhou2015cnnlocalization} & InceptionV3~\cite{szegedy2016rethinking} & 53.71 & 41.81 & 37.32 & 26.7 & 8.2 \\
     SPG~\cite{zhang2018self} & InceptionV3~\cite{szegedy2016rethinking} & 51.40 & 40.00 & 35.31 & 30.3 & 9.9 \\
     ADL~\cite{Choe_2019_CVPR} & InceptionV3~\cite{szegedy2016rethinking} & 51.29 & -  & - & 27.2 & - \\
     \rowcolor[gray]{0.9}
     $I^2C$-Ours & \textcolor{black}{InceptionV3}~\cite{szegedy2016rethinking} & \textcolor{black}{\textbf{46.89}} & \textcolor{black}{\textbf{35.87}} & \textcolor{black}{\textbf{31.50}} & \textcolor{black}{26.7} & \textcolor{black}{8.4} \\
     \hline
    \hline
  \end{tabular}
  \caption{Comparison of the localization error rate on ILSVRC validation set. Classification error rates is also presented for reference. 
  }\label{tab-comp-ilsvrc}
\end{table*}

\textbf{ILSVRC}
Table~\ref{tab-comp-ilsvrc} compares the proposed method with various baselines on the ILSVRC validation set.
$I^2C$ surpasses all the baseline methods in Top-1 and Gt-known localization error.
$I^2C$ based on ResNet50 achieves the lowest error rate of 45.17\%, significantly surpassing ADL by a large margin of 6.30\%.
It is notable ADL uses a stronger backbone network,~\ie, ResNet50-SE.
$I^2C$ based on InceptionV3 significantly surpasses the current state-of-the-art method, ADL, by $4.40\%$.
The results based on VGG16 are also notably better than the currently reported results by  1.58\% in Top-1. 
Additionally, the classification errors of our $I^2C$ method are competitive with all the counterparts based on the same backbones.
As demonstrated in Section~\ref{exper-setup}, the Gt-known localization metric can reflect the pure localization ability regardless the affect from classification results.
The proposed method achieves the best localization ability among the existing methods.
The lowest Gt-known localization error of $I^2C$ is $31.50\%$, outperforming the SPG approach by 3.81\%.
The $I^2C$ model based on VGG16 achieves 36.10\% in Gt-known localization error, which is also better than the counterparts.

\begin{wraptable}{R}{0.46\textwidth}\setlength{\tabcolsep}{10pt}
  \centering
  \small
  \begin{tabular}{l|cc}
    \hline
    \hline
    Methods & Top-1 & Top-5 \\
    \hline
    CAM~\cite{zhou2015cnnlocalization} & 56.33 & 46.47 \\
    ACoL~\cite{zhang2018adversarial} & 54.08 & 43.49 \\
    SPG~\cite{zhang2018self} & 53.36 & 42.28 \\
    DANet~\cite{xue2019danet} & 47.48 & 38.04 \\
    ADL~\cite{Choe_2019_CVPR} & 46.96 & - \\
    \hline
    \rowcolor[gray]{0.9}
    $I^2C$-Ours & \textcolor{black}{\textbf{44.01}} & \textcolor{black}{\textbf{31.66}}   \\
    \hline
    \hline
  \end{tabular}
  \caption{Localization error on the CUB-200-2011 test set.
  $I^2C$ significantly surpasses all the baselines.
  }\label{comp-cub}
\end{wraptable}

\textbf{CUB} 
We implement the proposed method on the CUB-200-2011 dataset following the baseline methods,~\eg, ACoL~\cite{zhang2018adversarial}, SPG~\cite{zhang2018self} and CAM~\cite{zhou2015cnnlocalization}.
InceptionV3 is chosen as the backbone network following~\cite{zhang2018self,Choe_2019_CVPR}.
Table~\ref{comp-cub} compares our method with the baselines.
$I^2C$ surpasses all the baseline methods on both Top-1 and Top-5 metrics, yielding the accuracies of Top-1 44.01\% and Top-5 31.66\% and significantly outperforming the current reported state-of-the-art errors by 2.95\% in Top-1 and 6.38\% in Top-5.  
\begin{figure*}[!t]
  \centering
  \includegraphics[width=1.0\textwidth]{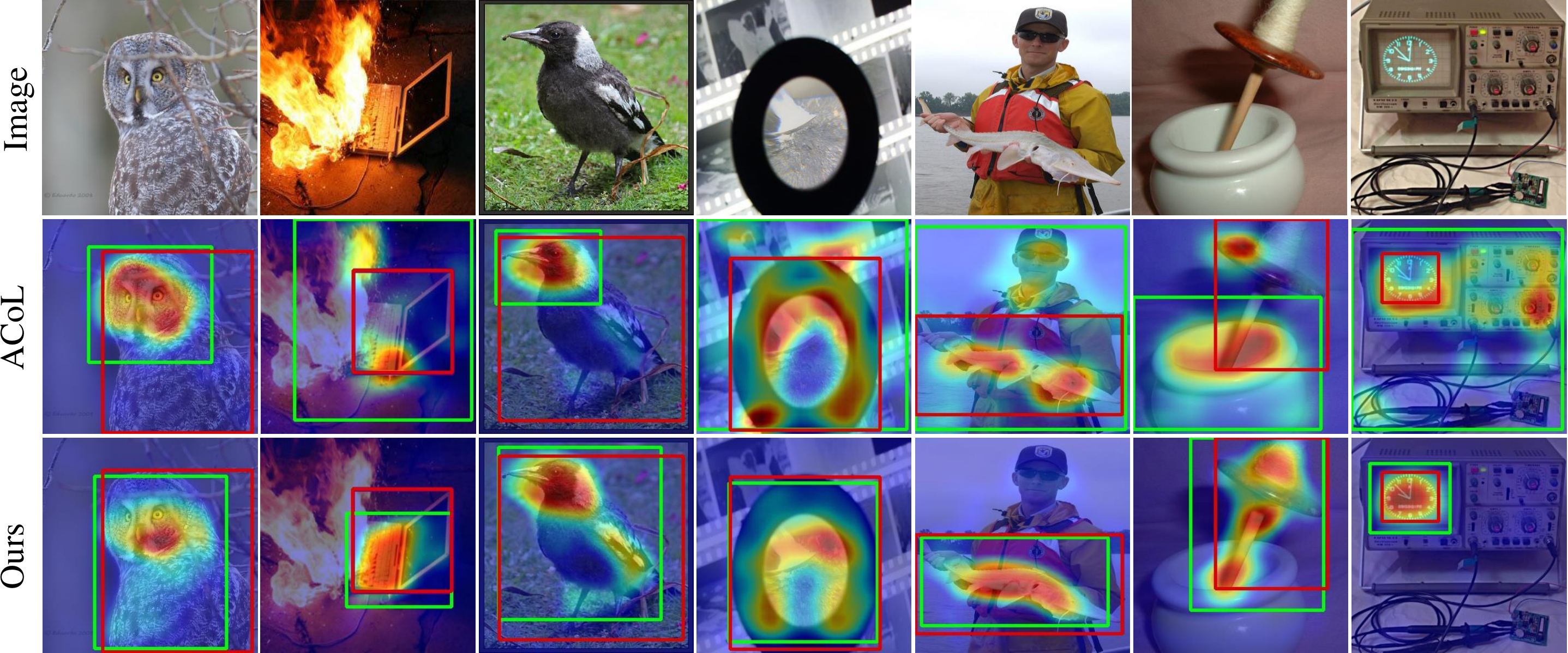}
  \caption{Comparison of the predicted bounding boxes with ACoL~\cite{zhang2018adversarial}. Our method obtains better localization maps and better bounding boxes. 
  \textit{The predicted boxes are in \textcolor{green}{green} and the ground-truth boxes are in \textcolor{red}{red}.}}\label{fig-comp-box}
\end{figure*}

\textbf{Visualization} 
In Figure~\ref{fig-comp-box}, we compare localization maps and the corresponding bounding boxes between the proposed method and ACoL~\cite{zhang2018adversarial}.
We can observe that localization maps produced by $I^2C$ can accurately highlight the object regions of interest.
The proposed method can not only reduce the noises from the irrelevant objects or stuff in the background regions,
but find more object-related regions accurately.
\begin{wraptable}{r}{0.4\textwidth}\setlength{\tabcolsep}{3pt}
  \centering
  \small
  \begin{tabular}{c|c|cc}
    \hline
    \hline
    DataSet & Resolution  & Gt-known \\
    \hline
    \multirow{2}{*}{ILSVRC} & $224\times224$  & 31.50 \\   
                            & $320\times320$  & 31.04 \\   
    \hline
    \multirow{2}{*}{CUB} & $224\times224$ & 27.40 \\
                        & $320\times320$ & 22.99 \\
    \hline
    \hline
  \end{tabular}
  \caption{Gt-known localization error with different input resolutions. 
  Enlarging input resolution slightly improve the localization on ILSVRC, while dramatically boost the localization performance on CUB.
  }\label{tab-ablation-resolution}
\end{wraptable}

\noindent As a result, it is easy to draw bounding boxes which can better match the target object regions as shown in Figure~\ref{fig-comp-box}.

\textit{In summary}, our method can successfully obtain better localization maps and accurate bounding boxes.
$I^2C$ surpasses all the baseline methods on both ILSVRC and CUB-200-2011.
We believe that there are two aspects accounting for the success of $I^2C$.
First, the proposed two constraints can retain pixel-level consistent between object features of different images, so that images can benefit from each other to obtain better localization maps.
Second, $I^2$ can not only increase the localization accuracy, but can get competitive classification results.

\subsection{Ablation study}
In this section, we analyse the insights and effectiveness of the different modules in the proposed method.
We conduct the ablation experiments based on InceptionV3. 
The experiments are on the most convincing large-scale dataset~\ie, ILSVRC, with the input resolution of 320 by 320, unless specifically specified.
When testing the classification and localization results, we do \textit{NOT} apply the ten-crop operation to enhance classification accuracies for convenience. 

\textbf{Does input resolution affect the localization ability?}
Table~\ref{tab-ablation-resolution} depicts the Gt-known localization errors \wrt~ different resolutions of input images.
We exclude the influences of classification results by comparing the Gt-known localization errors.
We see that enlarging the resolutions can decrease the localization errors with other configuration unchanged.
For ILSVRC, the gain is slight of only 0.46\%,
Dislike ILSVRC, CUB is more sensitive to the resolution of input images and the Gt-known error drops significantly by 4.41\%.
\begin{wraptable}{Rt}{0.6\textwidth}\setlength{\tabcolsep}{5pt}
  \centering
  \small
  \begin{tabular}{c|ccc}
    \hline
    \hline
    Methods & Plain & SC & SC + GC \\
    \hline
    Loc. & 53.71* & 49.07 & \textbf{48.08} \\
    Gt-known Loc. & 37.32 & 31.63 & \textbf{31.04} \\
    \hline
    \hline
  \end{tabular}
  \caption{Localization error on ILSVRC validation set using different configurations of the proposed constraints. ( $*$ indicates the numbers obtained with the classification results using the ten-crop operation. ) }\label{tab-ablation}
\end{wraptable}

\textbf{Are SC and GC really effective?}
Table~\ref{tab-ablation} compares the localization errors of the proposed constraints to the plain version network.
When using only the cross-entropy loss, the localization and the Gt-known localization errors are $53.71\%$ and $37.32\%$, respectively.
After adding the SC constraint, the localization accuracy is significantly improved and the error rate drops to $49.07\%$. 
The Gt-known localization error also decreases to $31.63\%$ by a large margin of $5.69\%$.
Furthermore, the localization performance can be boosted with the GC constraint.
By adding the proposed two constraints simultaneously, the localization and Gt-known errors can finally be reduced to $48.45\%$ and $31.30\%$, respectively.
Moreover, to verify the superiority of the proposed updating strategy in GC, we conduct an experiment of updating the global memory matrix with the back-propagation process.
The localization accuracies of such an updating method obtains 49.03\% and 32.09\% in localization and Gt-known error rates, respectively. 
The back-propagation updating strategy is worse than the proposed updating strategy.
Moreover, although the global constraint introduces a vector for each class, these vectors only involve negligible computational resources.
In particular, the number of parameters for the backbone network is 27M (InceptionV3).
GC only increase 0.8M parameters by 2.9\%.
During the forward stage of training, the Flops is 21.42G. 
GC only gain 3K Flops which can be totally ignored.
SC does not involve extra Flops nor parameters.
During the testing phase, neither SC nor GC involve any extra Flops.

\textbf{Is $I^2C$ sensitive to $\lambda_1$, $\lambda_2$ and $K$?}
\bm{$\lambda_1$} controls the relative importance of SC in training.
In order to maximize the performance of the proposed method and study the robustness to $\lambda_1$, we test the localization accuracy \wrt various values of $\lambda_1$.
We remove the GC constraint and only add the SC constraint.
Figure~\ref{fig:param-lambda1} illustrates the classification error, localization error and Gt-known localization errors when $\lambda_1$ changes in $40$ times of scale from $0.002$ to $0.08$.
We obtain the best results of $29.24\%$, $49.07\%$ and $31.63\%$ and the worst results of $32.13\%$, $50.82\%$ and $34.44\%$ in terms of the classification, localization and Gt-known localization error, respectively.
In general, with the increase of $\lambda_1$, the classification ability is getting better while Gt. localization is getting worse.
The localization errors with respect to the predicted labels achieve the best value of $49.07\%$ at $0.008$, because only the generated bounding boxes are counted as correct hits when the predicted labels of classification meet the ground-truth labels in this criterion.
It is notable that the gap between the best and worst values of the localization error is \textit{only $1.75\%$ over the changes of $40$ times in $\lambda_1$}, which means the proposed method is quite robust to the values of $\lambda_1$.

\bm{$\lambda_2$} controls the relative importance of GC in training.
According to Figure~\ref{fig:param-lambda1}, we leverage the best value of $\lambda_1=0.008$ to study the performance with $\lambda_2$ changing from $0.0001$ to $0.1$.
Figure~\ref{fig:param-lambda2} shows the accuracies of the proposed model adding both SC and GC.
The classification errors remain stable at around $30\%$ after applying GC.
The localization accuracy achieve lower error rates of $48.07\%$ compared to only using SC, which reflects the effectiveness of the proposed global consistency strategy.
As for the Gt-known localization metric,  the error rates decrease to $31.04\%$ by adding GC.
It is also notable that the model is robust to the change of the hyper-parameter $\lambda_2$ in \textit{a large range of 1,000 times from $0.0001$ to $0.1$}.
The localization error \textit{only fluctuates within a range of 0.96\%}.

\begin{figure}[t]
  \centering
  \begin{subfigure}[b]{0.30\linewidth}
    {\includegraphics[width=1.0\textwidth]{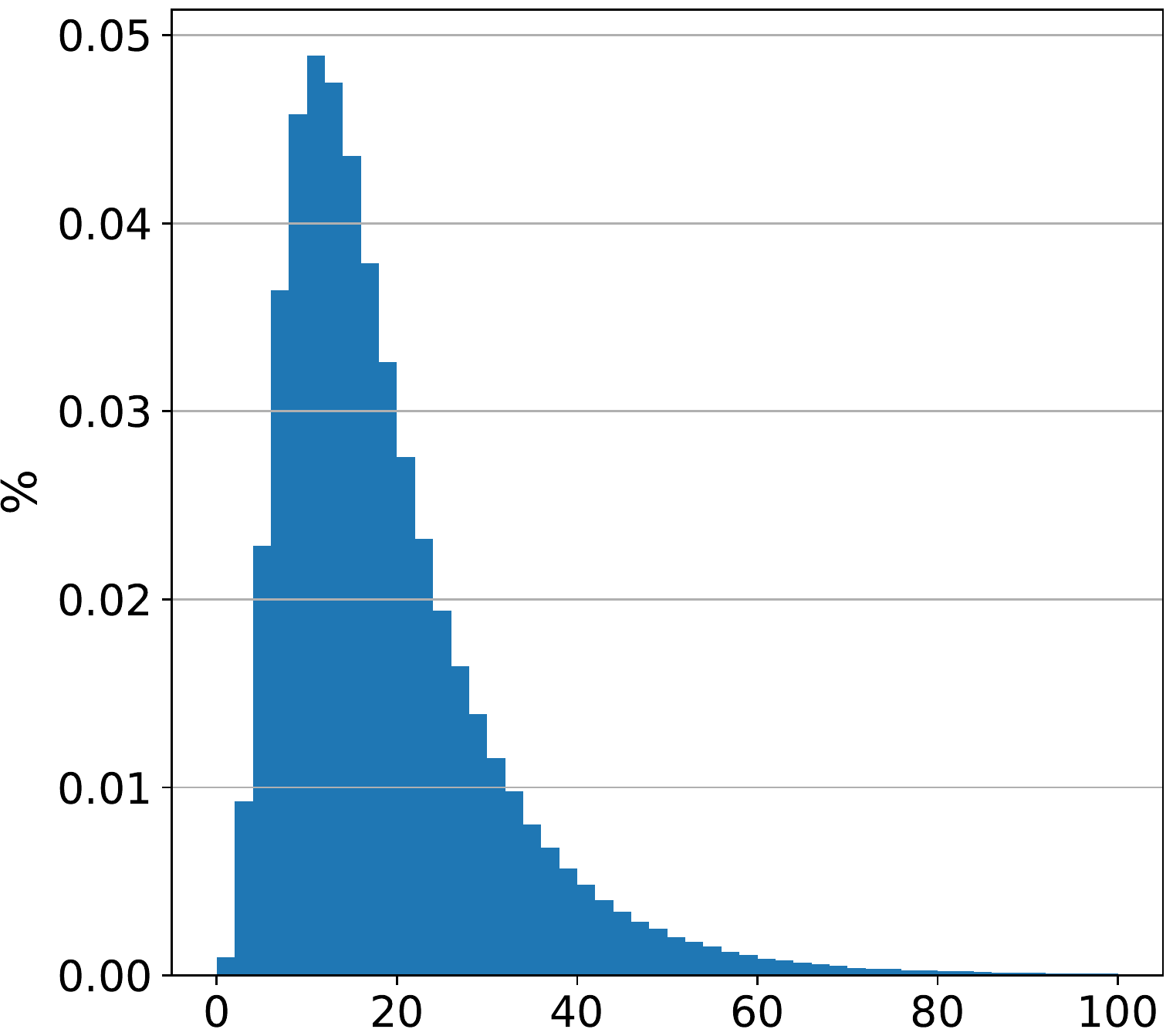}}
    \caption{}
  \end{subfigure}
  \hspace*{10pt}
  \begin{subfigure}[b]{0.4\linewidth}
    \includegraphics[width=1.0\textwidth]{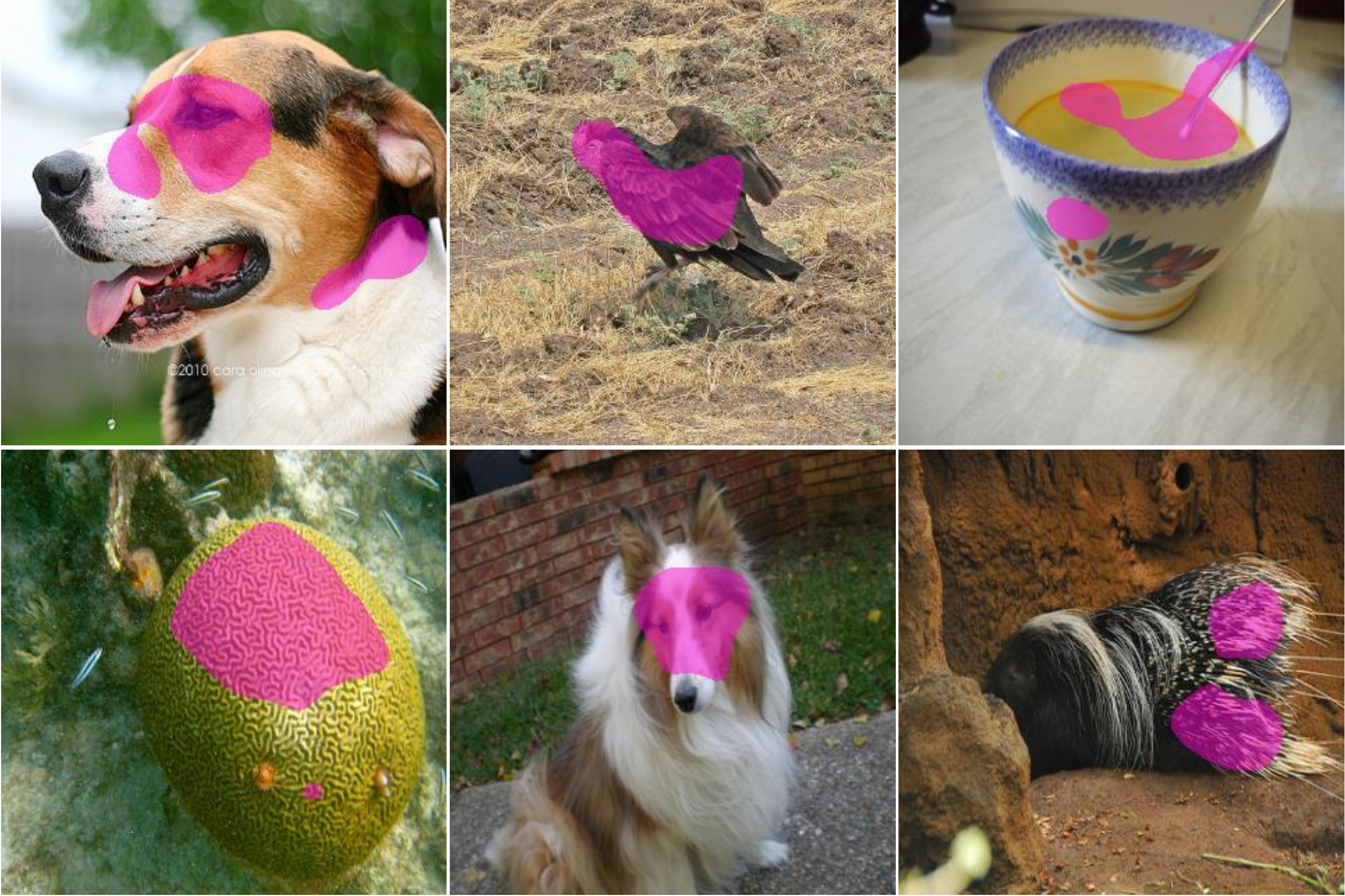}
    \caption{}
  \end{subfigure}
  \caption{\textit{(a)}: histogram of the number of object pixels with the threshold of $0.7$ on the ILSVRC training set.
  \textit{(b)}: identified object regions (in \textcolor{magenta}{magenta}) according to the localization maps}\label{fig-seed-num}
\end{figure}
\bm{$K$} is the number of the chosen object seeds.
For an input image with the resolution of $320$ by $320$, the networks downsample the resolution by a factor of $8$ and obtain the corresponding localization maps of $40$ by $40$ with $1600$ pixels.
Following the recommended threshold in SPG~\cite{zhang2018self}, we set the threshold $\delta$ to $0.7$ for discovering the object regions in each image. 
Figure~\ref{fig-seed-num} illustrates the histogram of the number of the identified object pixels on the ILSVRC training set.
We observe that most of the images contain object pixels in the range from 1 to 60.
Particularly, we choose $K=\{3, 20, 40, 60, 80, 100\}$ to study the performance changes with the variant number of sampled representative pixels in each image.
Figure~\ref{fig:param-K} shows the classification and localization accuracies with different $K$ values.
The classification error is lower when the number of sampled pixels is relatively larger.
The classification error rate reaches the lowest point of $28.40\%$ at $K=60$.
On the contrary, the localization error increases when we adopt a larger value of $K$, and the localization error is lowest at $K=3$.
\begin{wrapfigure}{R}{0.5\textwidth}
     \begin{subfigure}[t]{0.5\textwidth}
         \begin{minipage}{1.0\textwidth}
         \includegraphics[width=0.3\textwidth]{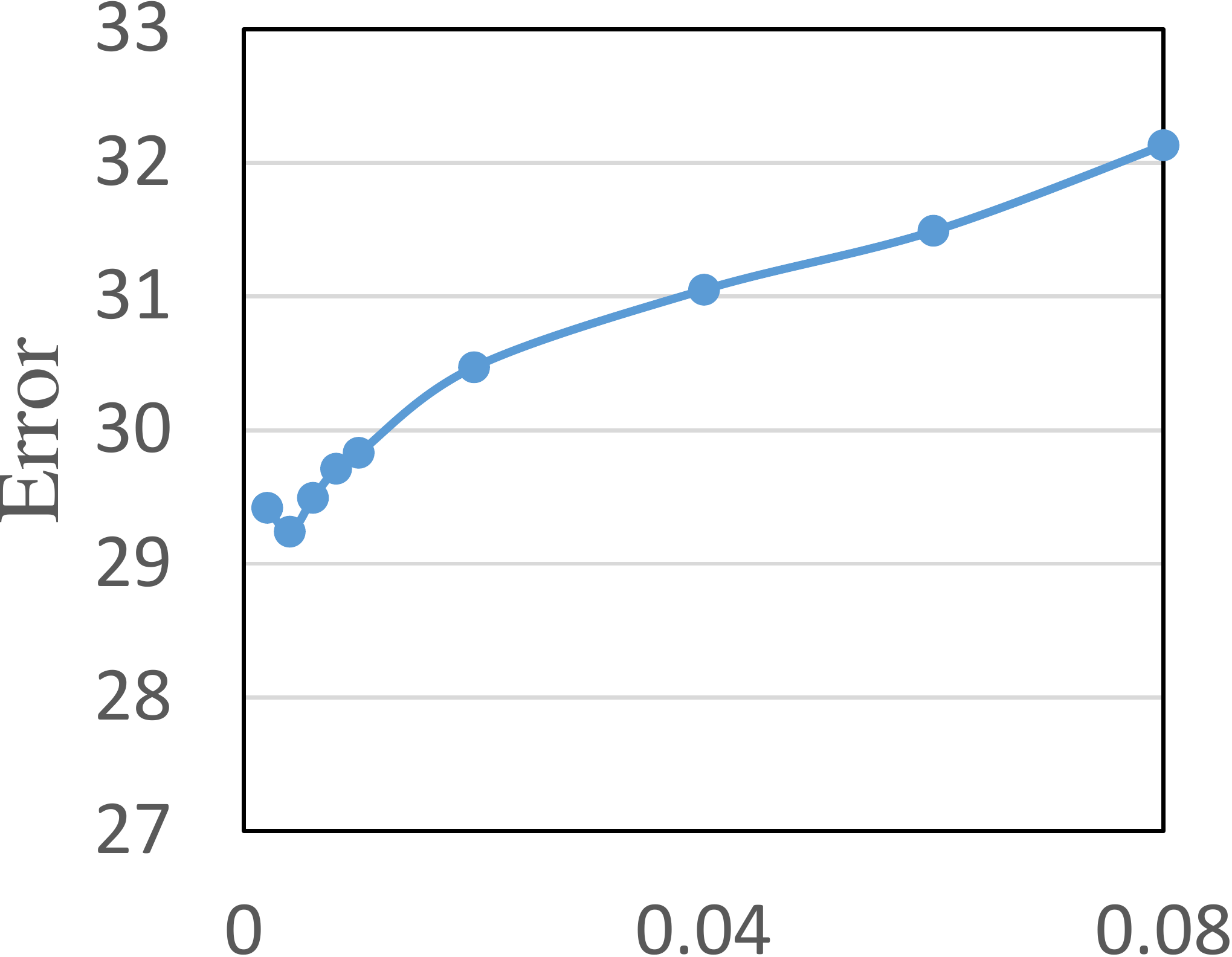}
         \includegraphics[width=0.3\textwidth]{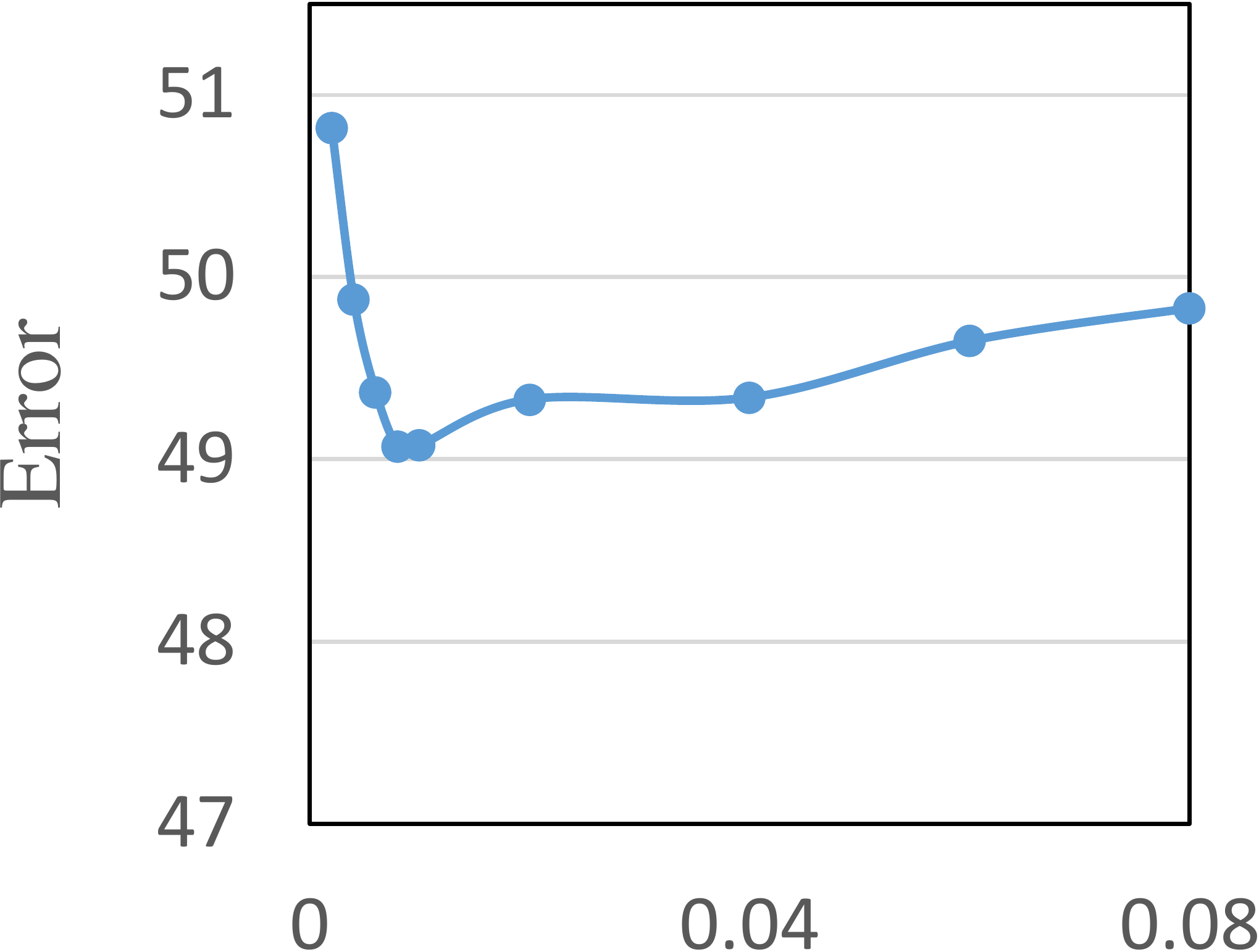}
         \includegraphics[width=0.3\textwidth]{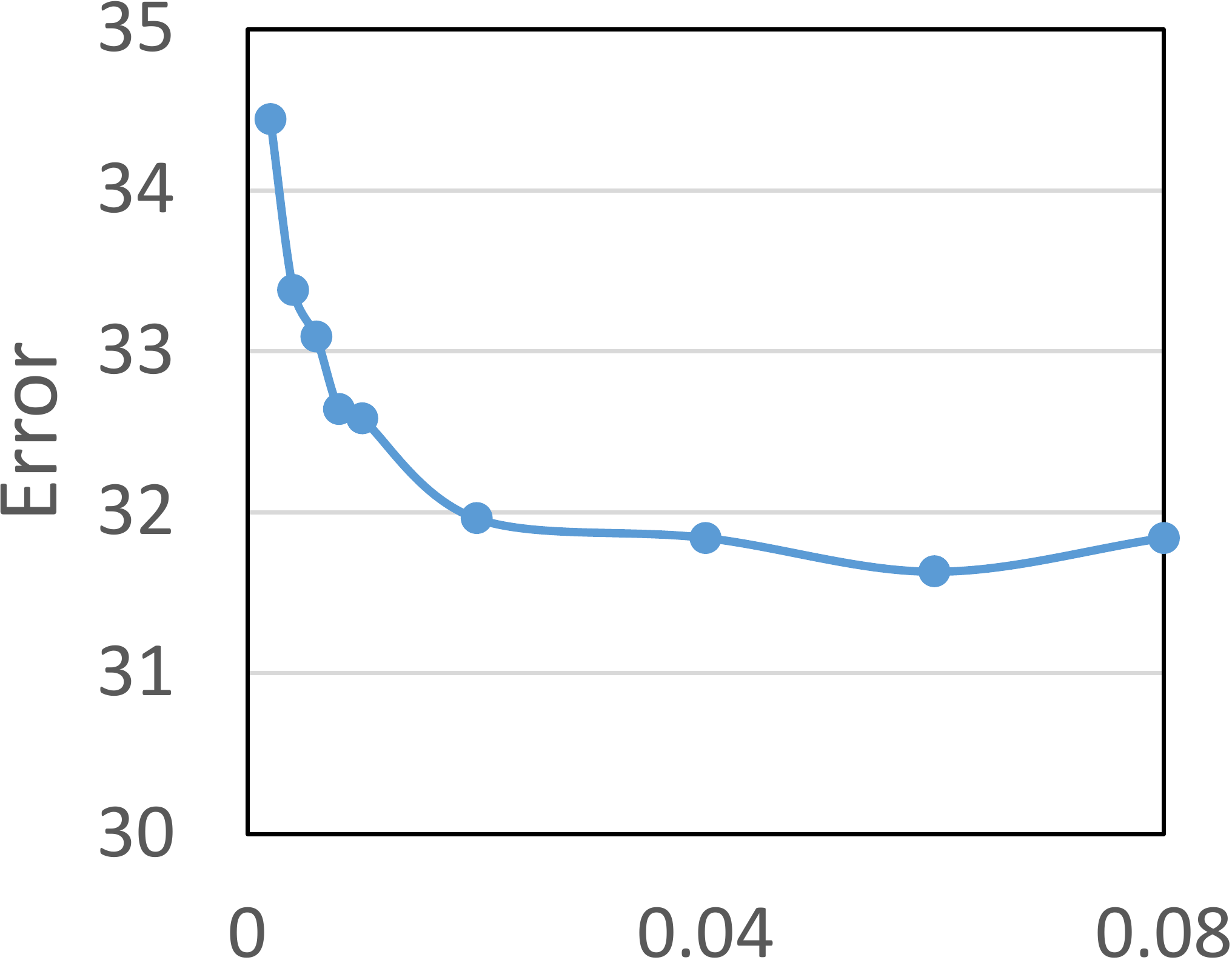}
         \end{minipage}
         \caption{$\lambda_1$}
         \label{fig:param-lambda1}
     \end{subfigure}
     \begin{subfigure}[t]{0.5\textwidth}
         \begin{minipage}{1.0\textwidth}
         \includegraphics[width=0.3\textwidth]{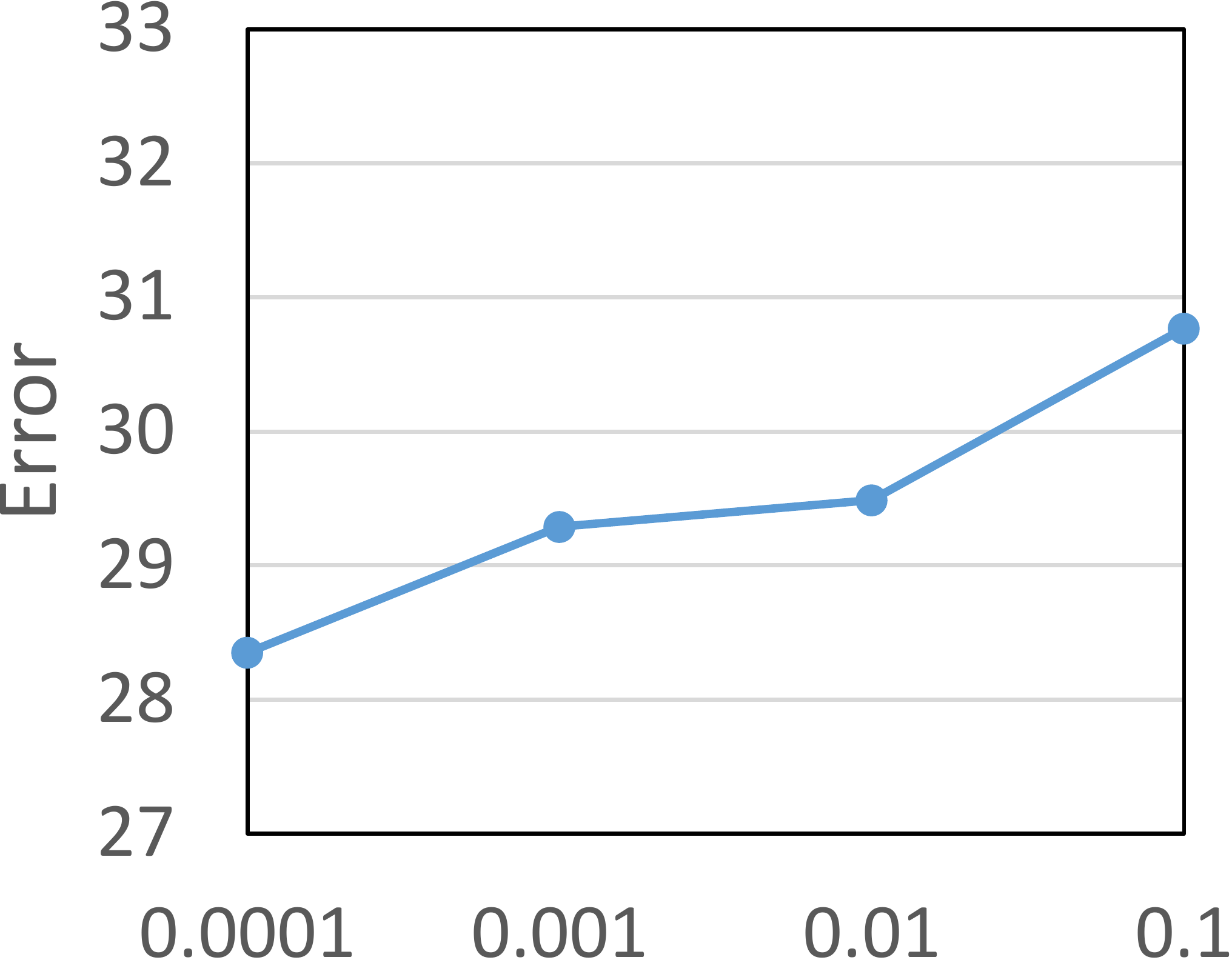}
         \includegraphics[width=0.3\textwidth]{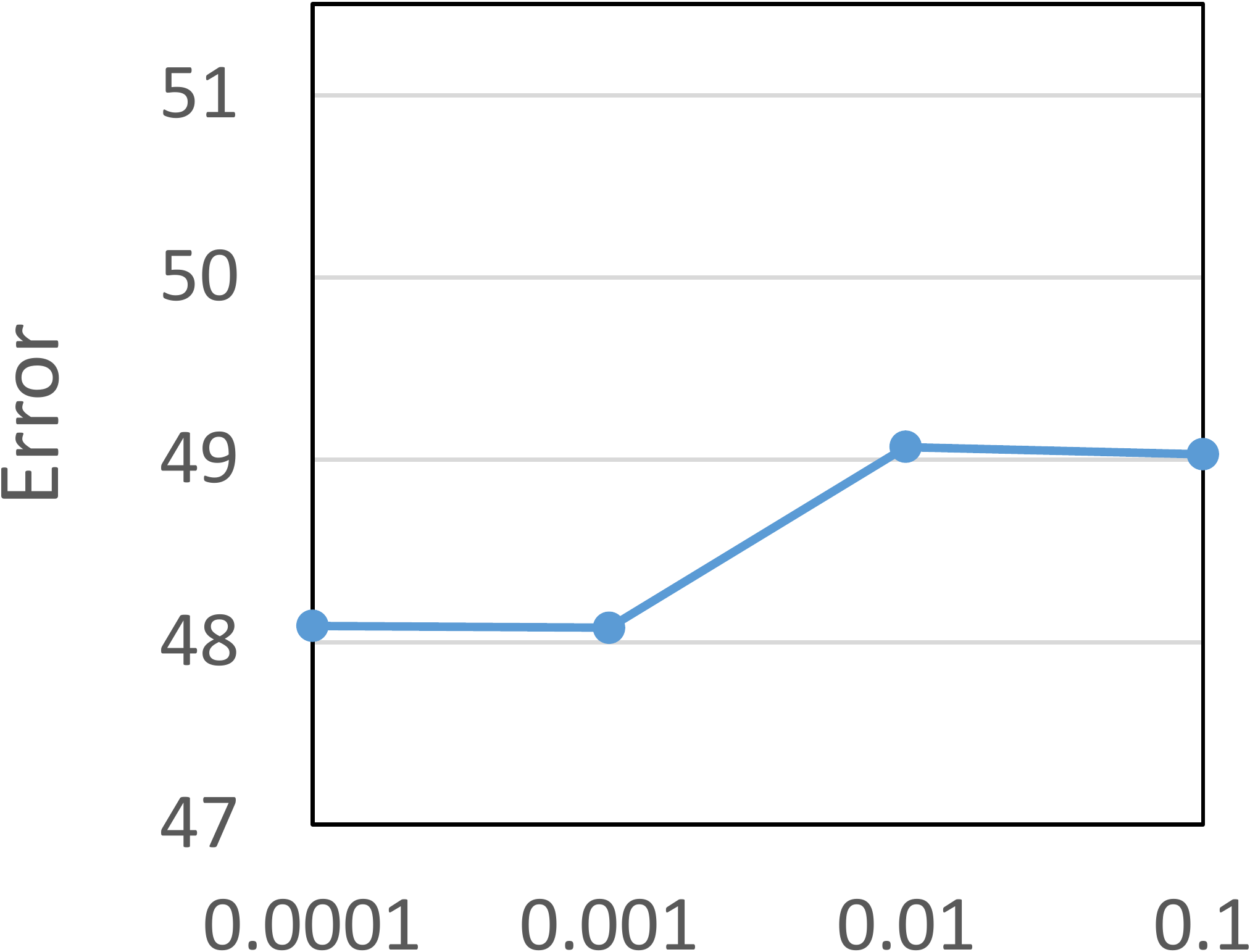}
         \includegraphics[width=0.3\textwidth]{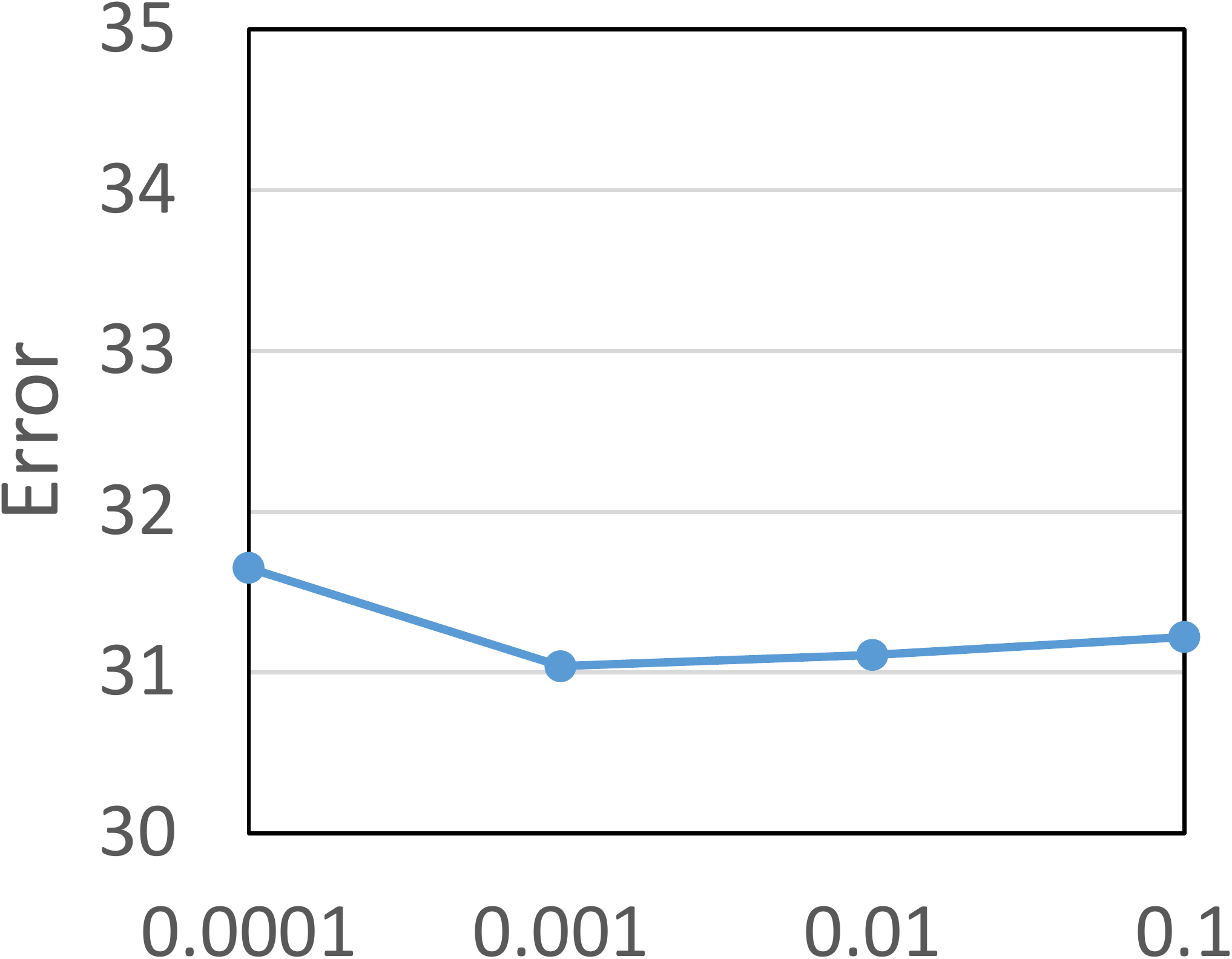}
         \end{minipage}
         \caption{$\lambda_2$}
         \label{fig:param-lambda2}
     \end{subfigure}
     \begin{subfigure}[t]{0.5\textwidth}
         \begin{minipage}{1.0\textwidth}
         \includegraphics[width=0.3\textwidth]{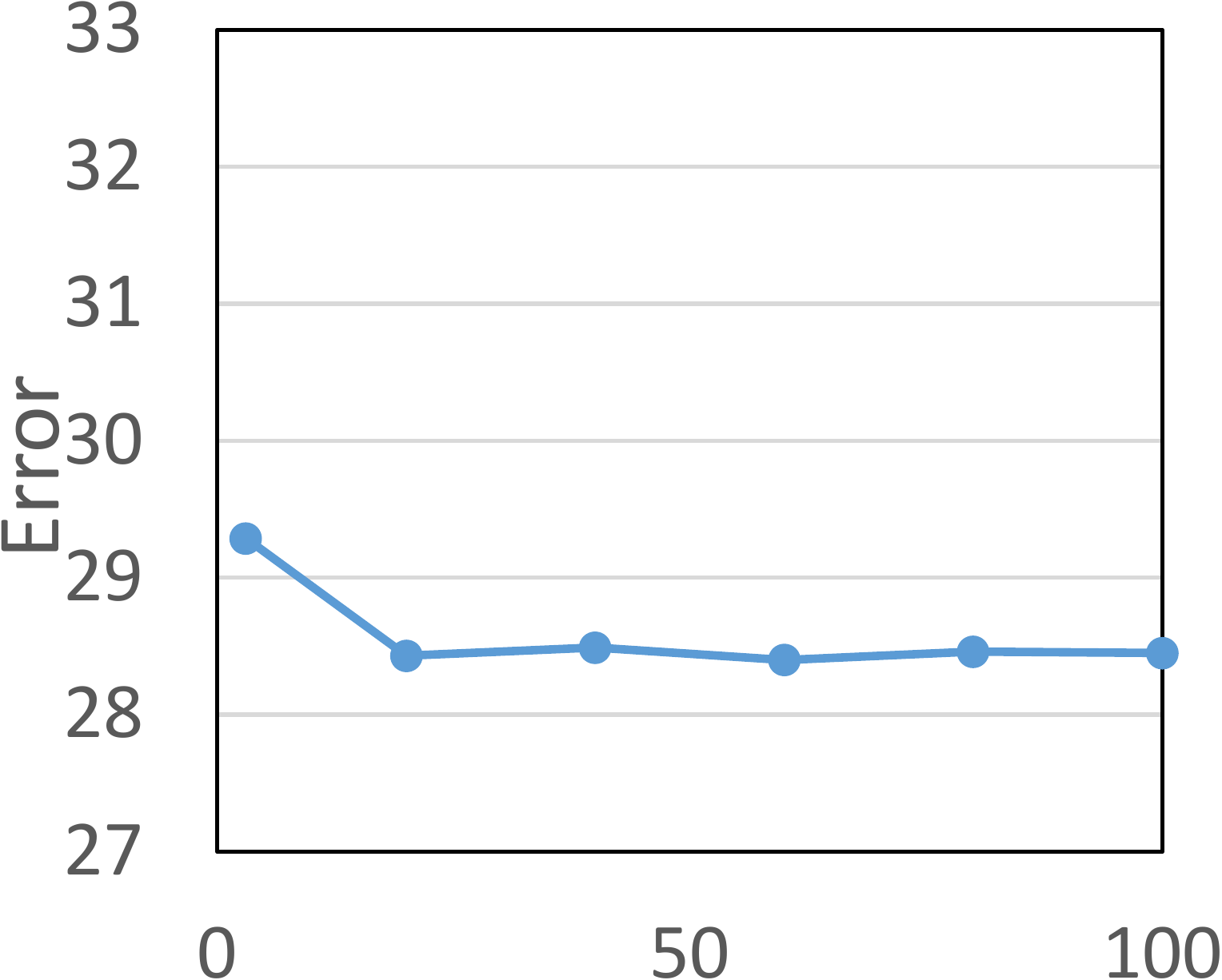}
         \includegraphics[width=0.3\textwidth]{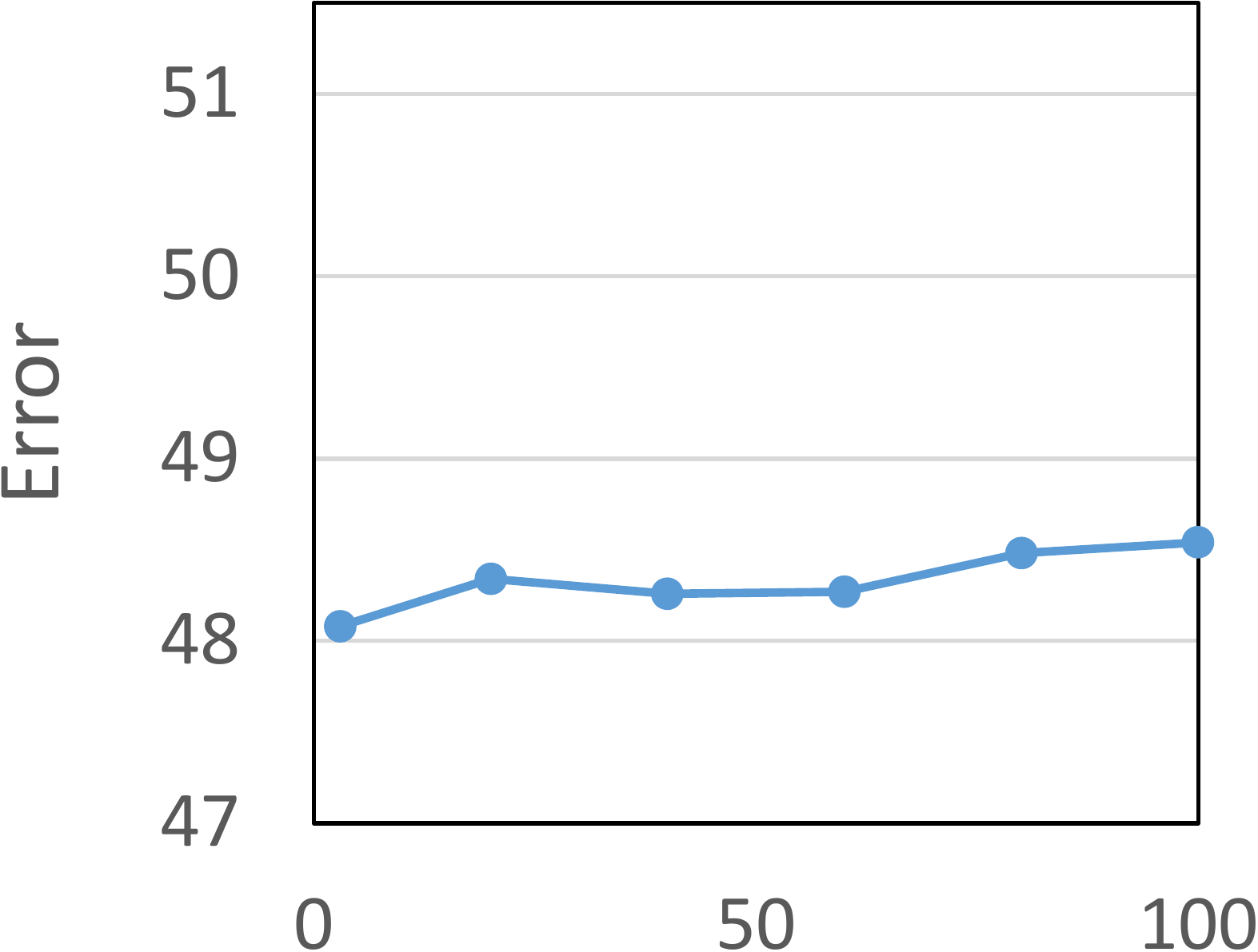}
         \includegraphics[width=0.3\textwidth]{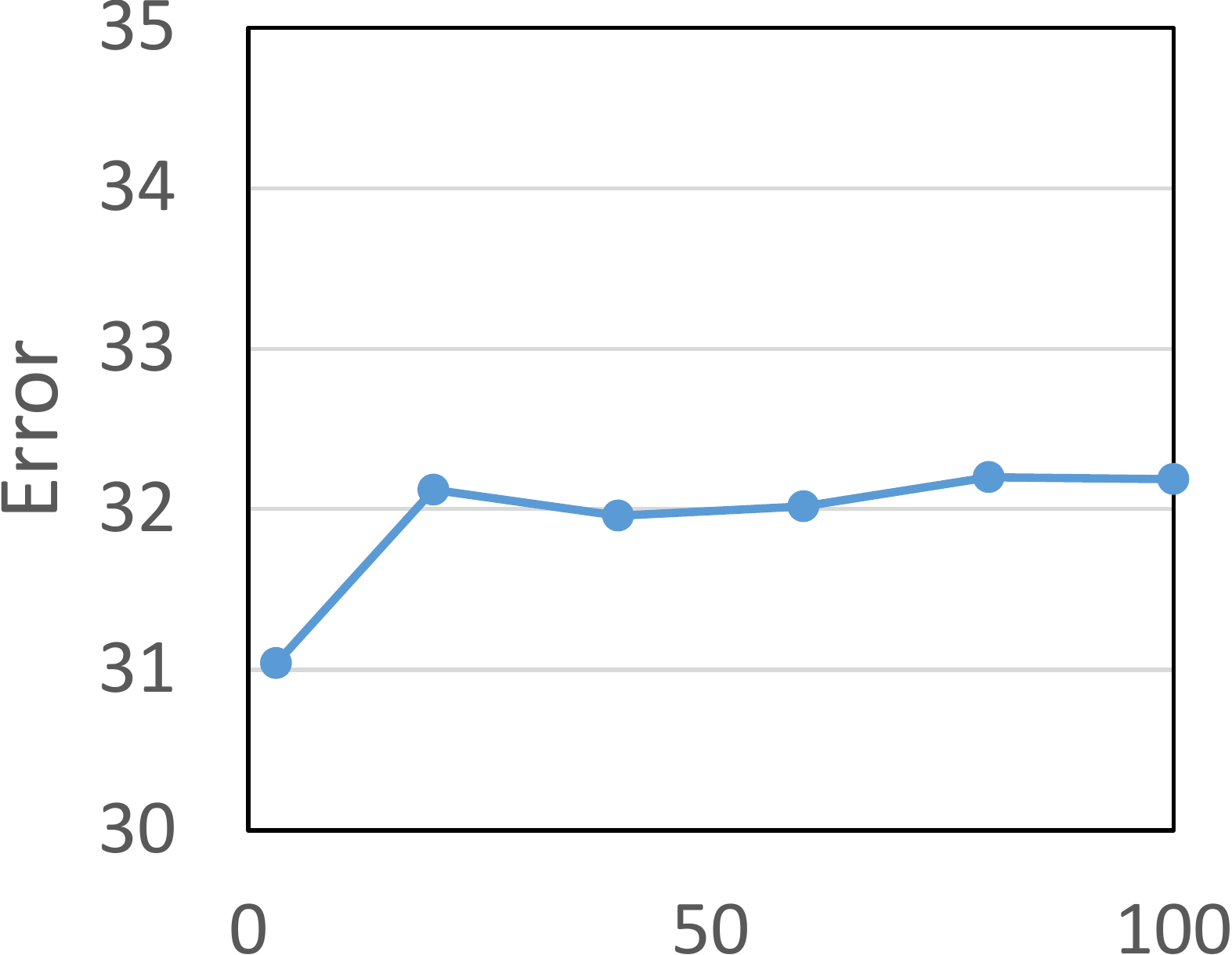}
         \end{minipage}
         \caption{$K$}
         \label{fig:param-K}
     \end{subfigure}
        \caption{Classification (\textit{left}), localization (\textit{middle}) and Gt-known Loc (\textit{right}) error rates with the changes of hyper-parameters, \ie, $\lambda_1$, $\lambda_2$ and $K$.}
        \label{fig:param}
\end{wrapfigure}
We believe the reason for this phenomenon is that the larger number may include more noises in the sampled object pixels, and more sampled pixels implicitly make the constraints stronger.
The proposed $I^2C$ is robust to the change of seed numbers, and \textit{the localization accuracy only fluctuates within a small range of 0.46\% with $K$ changing of 33 times from 3 to 100}.

In total, we can summarize that the improvement of the proposed method mainly attribute to three aspects.
First, SC can semantically drive the consistency between the object features in a batch, which improves the localization maps.
Second, GC can further drive the consistency throughout the entire dataset by forcing the object features towards their class-specific centers.
Third, the proposed updating strategy can effectively learn the class-specific centers with the training process.
Also, we have studied the robustness of the hyper-parameters.
$\lambda_1$ and $\lambda_2$ are for balancing the costs of SC and GC.
The experiment results show that the network performs superiorly in a very large range.
In addition, according to the changes of the localization performance \wrt the change of $K$. 
We can obtain satisfied accuracies just selecting a small value of $K$,~\eg, $K=3$.

\section{Conclusion}
We propose an Inter-Image Communication approach ($I^2C$) to improve the accuracy of localization maps through training classification networks.
First, we randomly select several object seeds according to the activated area of localization maps. These seed points are further employed to extract representation vectors of class-specific objects.
Second, the extracted vectors are leveraged to communicate between different objects of the same classes.
Concretely, stochastic consistency (SC) is proposed to optimize the objects within a mini-batch.
Global consistency (GC) is designed to keep consistency across minibatchs.
Also, a strategy is applied to update the global memory matrix.
Finally, the proposed $I^2C$ approach achieves the Top-1 localization error rate of $45.17\%$ on the ILSVRC validation set, surpassing the current state-of-the-art method.

\noindent \textbf{Acknowledgement.}\\
This work is partially supported by ARC DECRA DE190101315 and ARC DP200100938.
Xiaolin Zhang (No. 201606180026) is partially supported by the Chinese Scholarship Council.

\clearpage
%
%
\bibliographystyle{splncs04}
\bibliography{ref}

\begin{thebibliography}{10}
\providecommand{\url}[1]{\texttt{#1}}
\providecommand{\urlprefix}{URL }
\providecommand{\doi}[1]{https://doi.org/#1}

\bibitem{ahn2019weakly}
Ahn, J., Cho, S., Kwak, S.: Weakly supervised learning of instance segmentation
  with inter-pixel relations. In: {IEEE CVPR}. pp. 2209--2218 (2019)

\bibitem{ahn2018learning}
Ahn, J., Kwak, S.: Learning pixel-level semantic affinity with image-level
  supervision for weakly supervised semantic segmentation. In: {IEEE CVPR}. pp.
  4981--4990 (2018)

\bibitem{chen2014semantic}
Chen, L.C., Papandreou, G., Kokkinos, I., Murphy, K., Yuille, A.L.: Semantic
  image segmentation with deep convolutional nets and fully connected crfs.
  preprint arXiv:1412.7062  (2014)

\bibitem{chen2017deeplab}
Chen, L.C., Papandreou, G., Kokkinos, I., Murphy, K., Yuille, A.L.: Deeplab:
  Semantic image segmentation with deep convolutional nets, atrous convolution,
  and fully connected crfs. {IEEE TPAMI} pp. 834--848 (2017)

\bibitem{chen2018encoder}
Chen, L.C., Zhu, Y., Papandreou, G., Schroff, F., Adam, H.: Encoder-decoder
  with atrous separable convolution for semantic image segmentation. arXiv
  preprint arXiv:1802.02611  (2018)

\bibitem{cheng2019spgnet}
Cheng, B., Chen, L.C., Wei, Y., Zhu, Y., Huang, Z., Xiong, J., Huang, T.S.,
  Hwu, W.M., Shi, H.: Spgnet: Semantic prediction guidance for scene parsing.
  In: {IEEE ICCV}. pp. 5218--5228 (2019)

\bibitem{choe2020evaluating}
Choe, J., Oh, S.J., Lee, S., Chun, S., Akata, Z., Shim, H.: Evaluating weakly
  supervised object localization methods right. In: Proceedings of the IEEE/CVF
  Conference on Computer Vision and Pattern Recognition. pp. 3133--3142 (2020)

\bibitem{Choe_2019_CVPR}
Choe, J., Shim, H.: Attention-based dropout layer for weakly supervised object
  localization. In: {IEEE CVPR} (June 2019)

\bibitem{2009-imagenet}
Deng, J., Dong, W., Socher, R., Li, L.J., Li, K., Fei-Fei, L.: Imagenet: A
  large-scale hierarchical image database. In: {IEEE CVPR}. pp. 248--255 (2009)

\bibitem{duchi2011adaptive}
Duchi, J., Hazan, E., Singer, Y.: Adaptive subgradient methods for online
  learning and stochastic optimization. JMLR  \textbf{12}(Jul),  2121--2159
  (2011)

\bibitem{durand2017wildcat}
Durand, T., Mordan, T., Thome, N., Cord, M.: Wildcat: Weakly supervised
  learning of deep convnets for image classification, pointwise localization
  and segmentation. In: {IEEE CVPR}. pp. 642--651 (2017)

\bibitem{Fan_2020_CVPR}
Fan, J., Zhang, Z., Song, C., Tan, T.: Learning integral objects with
  intra-class discriminator for weakly-supervised semantic segmentation. In:
  {IEEE CVPR} (June 2020)

\bibitem{girshick15fastrcnn}
Girshick, R.: Fast r-cnn. In: arXiv preprint arXiv:1504.08083 (2015)

\bibitem{graves2013generating}
Graves, A.: Generating sequences with recurrent neural networks. arXiv preprint
  arXiv:1308.0850  (2013)

\bibitem{he2016deep}
He, K., Zhang, X., Ren, S., Sun, J.: Deep residual learning for image
  recognition. In: {IEEE CVPR}. pp. 770--778 (2016)

\bibitem{hou2018self}
Hou, Q., Jiang, P., Wei, Y., Cheng, M.M.: Self-erasing network for integral
  object attention. In: {NIPS}. pp. 549--559 (2018)

\bibitem{hu2018squeeze}
Hu, J., Shen, L., Sun, G.: Squeeze-and-excitation networks. In: {IEEE CVPR}.
  pp. 7132--7141 (2018)

\bibitem{huang2020ccnet}
Huang, Z., Wang, X., Wei, Y., Huang, L., Shi, H., Liu, W., Huang, T.: Ccnet:
  Criss-cross attention for semantic segmentation. {IEEE TPAMI}  (2020)

\bibitem{jiang2019integral}
Jiang, P.T., Hou, Q., Cao, Y., Cheng, M.M., Wei, Y., Xiong, H.K.: Integral
  object mining via online attention accumulation. In: {IEEE ICCV}. pp.
  2070--2079 (2019)

\bibitem{kingma2014adam}
Kingma, D.P., Ba, J.: Adam: A method for stochastic optimization. arXiv
  preprint arXiv:1412.6980  (2014)

\bibitem{kolesnikov2016seed}
Kolesnikov, A., Lampert, C.H.: Seed, expand and constrain: Three principles for
  weakly-supervised image segmentation. In: {ECCV}. pp. 695--711 (2016)

\bibitem{krizhevsky2012imagenet}
Krizhevsky, A., Sutskever, I., Hinton, G.E.: Imagenet classification with deep
  convolutional neural networks. In: {NIPS}. pp. 1097--1105 (2012)

\bibitem{liang2015towards}
Liang, X., Liu, S., Wei, Y., Liu, L., Lin, L., Yan, S.: Towards computational
  baby learning: A weakly-supervised approach for object detection. In: {IEEE
  ICCV}. pp. 999--1007 (2015)

\bibitem{oquab2015object}
Oquab, M., Bottou, L., Laptev, I., Sivic, J.: Is object localization for
  free?-weakly-supervised learning with convolutional neural networks. In:
  {IEEE CVPR}. pp. 685--694 (2015)

\bibitem{2015-papandreou-weakly}
Papandreou, G., Chen, L.C., Murphy, K., Yuille, A.L.: Weakly-and
  semi-supervised learning of a dcnn for semantic image segmentation. {IEEE
  ICCV}  (2015)

\bibitem{ren2015faster}
Ren, S., He, K., Girshick, R., Sun, J.: Faster r-cnn: Towards real-time object
  detection with region proposal networks. In: {NIPS}. pp. 91--99 (2015)

\bibitem{ILSVRC15}
Russakovsky, O., Deng, J., Su, H., Krause, J., Satheesh, S., Ma, S., Huang, Z.,
  Karpathy, A., Khosla, A., Bernstein, M., Berg, A.C., Fei-Fei, L.: {ImageNet
  Large Scale Visual Recognition Challenge}. {IJCV}  \textbf{115}(3),  211--252
  (2015). \doi{10.1007/s11263-015-0816-y}

\bibitem{shimoda2016distinct}
Shimoda, W., Yanai, K.: Distinct class-specific saliency maps for weakly
  supervised semantic segmentation. In: {ECCV}. pp. 218--234 (2016)

\bibitem{simonyan2013deep}
Simonyan, K., Vedaldi, A., Zisserman, A.: Deep inside convolutional networks:
  Visualising image classification models and saliency maps. arXiv preprint
  arXiv:1312.6034  (2013)

\bibitem{simonyan2014very}
Simonyan, K., Zisserman, A.: Very deep convolutional networks for large-scale
  image recognition. ICLR  (2015)

\bibitem{singh2017hide}
Singh, K.K., Lee, Y.J.: Hide-and-seek: Forcing a network to be meticulous for
  weakly-supervised object and action localization. arXiv preprint
  arXiv:1704.04232  (2017)

\bibitem{szegedy2015going}
Szegedy, C., Liu, W., Jia, Y., Sermanet, P., Reed, S., Anguelov, D., Erhan, D.,
  Vanhoucke, V., Rabinovich, A.: Going deeper with convolutions. In: {IEEE
  CVPR}. pp.~1--9 (2015)

\bibitem{szegedy2016rethinking}
Szegedy, C., Vanhoucke, V., Ioffe, S., Shlens, J., Wojna, Z.: Rethinking the
  inception architecture for computer vision. In: {IEEE CVPR}. pp. 2818--2826
  (2016)

\bibitem{WahCUB_200_2011}
Wah, C., Branson, S., Welinder, P., Perona, P., Belongie, S.: {The Caltech-UCSD
  Birds-200-2011 Dataset}. Tech. Rep. CNS-TR-2011-001, California Institute of
  Technology (2011)

\bibitem{Wang_2020_CVPR}
Wang, Y., Zhang, J., Kan, M., Shan, S., Chen, X.: Self-supervised equivariant
  attention mechanism for weakly supervised semantic segmentation. In: IEEE/CVF
  Conference on Computer Vision and Pattern Recognition (CVPR) (June 2020)

\bibitem{wei2017object}
Wei, Y., Feng, J., Liang, X., Cheng, M.M., Zhao, Y., Yan, S.: Object region
  mining with adversarial erasing: A simple classification to semantic
  segmentation approach. In: {IEEE CVPR} (2017)

\bibitem{wei2016learning}
Wei, Y., Liang, X., Chen, Y., Jie, Z., Xiao, Y., Zhao, Y., Yan, S.: Learning to
  segment with image-level annotations. PR  (2016)

\bibitem{wei2015stc}
Wei, Y., Liang, X., Chen, Y., Shen, X., Cheng, M.M., Feng, J., Zhao, Y., Yan,
  S.: Stc: A simple to complex framework for weakly-supervised semantic
  segmentation. {IEEE TPAMI}  (2016)

\bibitem{wei2018revisiting}
Wei, Y., Xiao, H., Shi, H., Jie, Z., Feng, J., Huang, T.S.: Revisiting dilated
  convolution: A simple approach for weakly-and semi-supervised semantic
  segmentation. In: {IEEE CVPR}. pp. 7268--7277 (2018)

\bibitem{xue2019danet}
Xue, H., Liu, C., Wan, F., Jiao, J., Ji, X., Ye, Q.: Danet: Divergent
  activation for weakly supervised object localization. In: {IEEE ICCV}. pp.
  6589--6598 (2019)

\bibitem{yun2019cutmix}
Yun, S., Han, D., Oh, S.J., Chun, S., Choe, J., Yoo, Y.: Cutmix: Regularization
  strategy to train strong classifiers with localizable features. In: {IEEE
  ICCV}. pp. 6023--6032 (2019)

\bibitem{zhang2018adversarial}
Zhang, X., Wei, Y., Feng, J., Yang, Y., Huang, T.: Adversarial complementary
  learning for weakly supervised object localization. In: IEEE CVPR (2018)

\bibitem{zhang2018self}
Zhang, X., Wei, Y., Kang, G., Yang, Y., Huang, T.: Self-produced guidance for
  weakly-supervised object localization. In: {ECCV}. Springer (2018)

\bibitem{zhang2020rethinking}
Zhang, X., Wei, Y., Yang, Y., Wu, F.: Rethinking localization map: Towards
  accurate object perception with self-enhancement maps. arXiv preprint
  arXiv:2006.05220  (2020)

\bibitem{zhao2017pyramid}
Zhao, H., Shi, J., Qi, X., Wang, X., Jia, J.: Pyramid scene parsing network.
  In: {IEEE CVPR}. pp. 2881--2890 (2017)

\bibitem{zhou2015cnnlocalization}
Zhou, B., Khosla, A., A., L., Oliva, A., Torralba, A.: {Learning Deep Features
  for Discriminative Localization.} {IEEE CVPR}  (2016)

\bibitem{zhou2018interpreting}
Zhou, B., Bau, D., Oliva, A., Torralba, A.: Interpreting deep visual
  representations via network dissection. {IEEE TPAMI}  (2018)

\bibitem{zhu2017soft}
Zhu, Y., Zhou, Y., Ye, Q., Qiu, Q., Jiao, J.: Soft proposal networks for weakly
  supervised object localization. arXiv preprint arXiv:1709.01829  (2017)

\end{thebibliography}
\end{document}